\documentclass[conference]{IEEEtran}

\newcommand{\cmmnt}[1]{}

\IEEEoverridecommandlockouts 
\usepackage{cite}
\usepackage{amsmath,amssymb,amsfonts}
\usepackage{algorithmic}
\usepackage{graphicx,color}
\usepackage{textcomp}

\usepackage{comment}
\usepackage{blindtext}
\usepackage{dirtytalk}
\usepackage{array}
\usepackage{authblk}
\usepackage{url}
\usepackage{graphics}
\usepackage{graphicx}
\usepackage{epsfig}
\usepackage{amssymb}
\usepackage{amsmath}
\usepackage{breqn}
\usepackage{adjustbox}
\usepackage[ruled,vlined]{algorithm2e}

\usepackage{mathtools}
\usepackage{lscape}
\usepackage{pdflscape}
\usepackage{longtable}
\usepackage{makecell}
\usepackage{lettrine}
\usepackage{caption}
\usepackage{supertabular,booktabs}
\usepackage{color}
\usepackage{soul}

\usepackage{amsmath}
\usepackage{graphicx}
\usepackage{subcaption}
\usepackage[colorlinks=true, allcolors=blue]{hyperref}
\usepackage{amssymb}
\usepackage[table,xcdraw]{xcolor}
\usepackage{tabularx} 
\usepackage{tikz}

\newcommand\copyrighttext{%
  \footnotesize \textcopyright \the\year{} IEEE. Personal use of this material is permitted.  Permission from IEEE must be obtained for all other uses, in any current or future media, including reprinting/republishing this material for advertising or promotional purposes, creating new collective works, for resale or redistribution to servers or lists, or reuse of any copyrighted component of this work in other works.}

\newcommand\copyrightnotice{%
\begin{tikzpicture}[remember picture,overlay]
\node[anchor=south,yshift=10pt] at (current page.south) {\fbox{\parbox{\dimexpr0.75\textwidth-\fboxsep-\fboxrule\relax}{\copyrighttext}}};
\end{tikzpicture}%
}

\newcolumntype{M}{>{$\displaystyle}c<{$}}
\newcommand\AddLabel[1]{\refstepcounter{equation}(\theequation)\label{#1}}
\newcolumntype{L}{>{\collectcell\AddLabel}r<{\endcollectcell}}

\title{\LARGE
Applied Federated Model Personalisation in the Industrial Domain: A Comparative Study}

\begin{document}

\author{Ilias Siniosoglou\IEEEauthorrefmark{1}, Vasileios Argyriou \IEEEauthorrefmark{2}, George Fragulis\IEEEauthorrefmark{1}, Panagiotis Fouliras\IEEEauthorrefmark{3}, \\ Georgios Th. Papadopoulos\IEEEauthorrefmark{4}, Anastasios Lytos\IEEEauthorrefmark{5} and Panagiotis Sarigiannidis\IEEEauthorrefmark{1}\IEEEauthorrefmark{6}

\thanks{\IEEEauthorrefmark{1} I. Siniosoglou, G. Fragulis and P. Sarigiannidis are with the Department of Electrical and Computer Engineering, University of Western Macedonia, Kozani, Greece - \texttt{E-Mail: \{isiniosoglou, gfragulis, psarigiannidis\}@uowm.gr}}

\thanks{\IEEEauthorrefmark{2} V. Argyriou is with the Department of Networks and Digital Media, Kingston University, Kingston upon Thames, United Kingdom - \texttt{E-Mail: vasileios.argyriou@kingston.ac.uk}}

\thanks{\IEEEauthorrefmark{3} P. Fouliras is with the Department of Applied Informatics, University of Macedonia, Thessaloniki, Greece -
\texttt{E-Mail: pfoul@uom.edu.gr}}

\thanks{\IEEEauthorrefmark{4} G. T. Papadopoulos is with the Department of Informatics and Telematics Harokopio University of Athens, Greece \texttt{E-Mail: G.th.papadopoulos@hua.gr}}

\thanks{\IEEEauthorrefmark{5} A. Lytos is with Sidroco Holdings Ltd., Nicosia, Cyprus -\texttt{E-Mail: alytos@sidroco.com}}

\thanks{\IEEEauthorrefmark{6} P. Sarigiannidis is with the R\&D Department, MetaMind Innovations P.C., Kozani, Greece - \texttt{E-Mail: psarigiannidis@metamind.gr}}

}

\maketitle
\copyrightnotice
\thispagestyle{empty}
\pagestyle{empty}


\begin{abstract}
The time-consuming nature of training and deploying complicated Machine and Deep Learning (DL) models for a variety of applications continues to pose significant challenges in the field of Machine Learning (ML). These challenges are particularly pronounced in the federated domain, where optimizing models for individual nodes poses significant difficulty. Many methods have been developed to tackle this problem, aiming to reduce training expenses and time while maintaining efficient optimisation.
Three suggested strategies to tackle this challenge include Active Learning, Knowledge Distillation, and Local Memorization. These methods enable the adoption of smaller models that require fewer computational resources and allow for model personalization with local insights, thereby improving the effectiveness of current models.
The present study delves into the fundamental principles of these three approaches and proposes an advanced Federated Learning System that utilises different Personalisation methods towards improving the accuracy of AI models and enhancing user
experience in real-time NG-IoT applications, investigating the efficacy of these techniques in the local and federated domain. 
The results of the original and optimised models are then compared in both local and federated contexts using a comparison analysis. The post-analysis shows encouraging outcomes when it comes to optimising and personalising the models with the suggested techniques.
\end{abstract}


\section{INTRODUCTION}
\label{Introduction}

In the past years, the utilization of intelligent devices has seen an exponential growth. Internet of Things (IoT) devices are being integrated for a multitude of purposes in areas such as smart grids, healthcare, smart buildings, and precision agriculture. These devices constantly produce a large amount of data that needs to be accurately processed and securely stored. Artificial Intelligence (AI) is a concept used to extract meaningful insights from raw IoT data. However, in order to successfully train a machine learning model, a large amount of annotated data is necessary. Furthermore, due to the large amount of data produced by the intelligent devices, the centralization of the data processing for the creation of machine learning models is no longer a viable option.

Federated learning is a machine learning setting where multiple entities (clients) collaborate in solving a machine learning problem, under the coordination of a central server or service provider. Each client’s raw data is stored locally and not exchanged or transferred; instead, focused updates intended for immediate aggregation are used to achieve the learning objective \cite{Zhao2023, Zhao2024}. Still, in most cases, as all other machine learning applications, federated learning requires a large amount of annotated data to complete a federated training session, where each client locally trains the model and sends it back to the server for fusion and global model generation. In addition, the global model produced after federated learning, although it is able to generalize, it is not customized to each client’s/intelligent device’s behaviour. As such, personalization methods are necessary to ensure that models produced after federated learning are customized to each client \cite{Qi2024}. Finally, personalization techniques should require less data to customize the global model, in order not to further consume large processing power from the constrained devices.

AI model personalization \cite{Kairouz2021} involves adapting an AI model to a specific user or group of users. The primary goal of AI model personalization is to improve the accuracy and relevance of AI models for users. Personalization is achieved by considering the user's historical data, preferences, and behaviour patterns. AI models are designed to learn from data, and personalization involves providing the AI model with personalized data that is relevant to the user.

AI model personalization is important for several reasons. Firstly, personalization improves the accuracy of AI models. When an AI model is personalized, it is more likely to provide accurate predictions or recommendations based on the user's preferences and usage patterns. Secondly, personalization enhances the user experience. Personalized AI models are more engaging and provide users with a sense of control over the content they receive. Lastly, personalization can lead to increased revenue for businesses. Personalized AI models can help businesses to improve customer satisfaction, retention, and loyalty.

The main focus of this paper is to present and evaluate an advanced Federated Learning System that utilises different Personalisation methods, such as Active Learning \cite{Pengzhen2020}, Knowledge Distillation \cite{Hinton2015DistillingTK} and Local Memorisation \cite{Li2019FedMDHF}, towards improving the accuracy of AI models and enhancing user experience in real-time Next-Generation Internet of Things (NG-IoT) applications, such as Smart Farming, Smart Home Energy Management, and Supply Chain Forecasting. This research looks at a variety of deep learning models, including the popular Long Short-Term Memory (LSTM) models \cite{Google2019}, the more recent Transformer models \cite{Yishay2020, Zhao20234}, and traditional models like simple Deep Neural Networks (DNN) and Linear Regression (LR). Through investigating these techniques on different kinds of models, we can have a more thorough grasp of the possible advantages and disadvantages of this approach for edge personalisation of federated learning models. These experiments also aim to shed light on the efficacy and constraints of these methods for enhancing and optimising pre-trained deep learning models, in addition to investigating their positive effects on standard deep learning models.

The overall contributions of this paper can be summarised as follows:

\begin{itemize}
    \item Proposes an advanced Federated Learning System that utilises different Personalisation methods towards improving the accuracy of AI models and enhancing user experience in real-time NG-IoT applications.
    \item Explores the advantages and limitations of different Personalisation methods in a Federated Learning Ecosystem.
    \item Investigates the application of Federated Learning and Personalisation to benchmark DL architectures.
    \item Provides a comparative study of a Personalised Federated Learning in different kinds of real-world decentralised datasets
\end{itemize}



The rest of this paper is organized as follows: the related work is discussed in Section \ref{Related work}, followed by an overview of the methodology in Section \ref{Methodology}. Section \ref{Evaluation} provides a comprehensive analysis of the available data, as well as a series of quantitative results. Section \ref{Conclusions} offers concluding remarks.


\section{RELATED WORK}
\label{Related work}
In the past years, the utilization of intelligent devices or systems has seen an exponential growth. IoT devices are being integrated for a multitude of purposes in areas such as smart grids, healthcare, smart buildings, and precision agriculture. These devices constantly produce a large amount of data that needs to be accurately processed and securely stored. Artificial Intelligence (AI) is a concept used to extract meaningful insights from raw IoT data. However, in order to successfully train a machine learning model, a large amount of annotated data is necessary. Furthermore, due to the large amount of data produced by the intelligent devices, the centralization of the data processing for the creation of machine learning models is no longer a viable option.

This is why federated learning has emerged. Federated learning is a machine learning methodology that involves the collaboration of several entities, known as clients, under the direction of a central server or service provider, in order to solve machine learning problems. To achieve the learning purpose, customised updates meant for instantaneous aggregation are used in place of each client's raw data, which is stored locally and never shared or transferred.

Still, federated learning requires a large amount of annotated data to complete a federated training session, where each client locally trains the model and sends it back to the server for fusion and global model generation. In addition, the global model produced after federated learning, although it is able to generalize, i.e., be able to predict a wider range of samples, it is not customized to each client’s/intelligent device’s behaviour. As such, personalization methods are necessary to ensure that models produced after federated learning are customized to each client. Finally, personalization techniques should require less data to customize the global model, in order to not further consume large processing power from the constrained devices.

\subsection{Active Learning}
Active learning is a machine learning technique that finds examples that are especially useful for learning, hence reducing the amount of labelled samples required for model training. Numerous research have investigated the use of this methodology in the identification of cyberattacks. 

Notably, network intrusion detection using active learning can be viewed as an unsupervised task \cite{Kloft2009}. The authors focus on exploring the way on how anomaly detection can be equipped with active learning. In particular, the authors suggest a novel querying approach that targets low-confidence data points in an effort to minimise labelling efforts. They use support vector domain description (SVDD) as the foundation for their anomaly detection method.
The authors focus on integrating the approach of active learning with SVDD in order to retrain the model after querying for data points, by utilizing unlabelled as well as newly labelled data. The outcomes of the experiments showed that the ActiveSVDDs reduced labelling work while effectively differentiating between attack and normal data.

The authors of \cite{Naoki2006} suggest a technique that uses artificially generated examples to represent outliers and turns outlier identification into a classification challenge. They then employ selective sampling with active learning in an effort to address issues like significant computing overhead and conclusions about outlier detection that are difficult to comprehend.
Specifically, the authors consider the application of ensemble-based minimum margin active learning, which is a combination of querying by committee and ensemble methodology for classification accuracy enhancement. Experiments show that the suggested methodology performs better than methods that use traditional classification procedures but apply comparable reduction strategies.

Regarding unsupervised anomaly detection tasks, the authors in \cite{pimentel2020deep} suggest combining active learning techniques with deep learning methods to differentiate outliers from regular data.
The authors propose active anomaly detection as an alternative approach to traditional unsupervised anomaly detection procedures, due to the latter one’s difficulty of separating anomalous instances from normal samples. In active anomaly detection, feedback can be given by experts in order to point to anomalous examples in the dataset, thus providing valuable input to the model \cite{Li2023}. An Unsupervised to Active Inference (UAI) layer is added to unsupervised deep learning systems in order to achieve this. Specifically, at each training step, the most probably anomalous data points are selected through the most-likely positive querying strategy and sent to be labelled by the experts before the actual training begins. The outcomes of the experiment showed that the models' performance was either the same or better than that of their peers who did not apply active learning strategies.

\subsection{Local Memorization}

Local memorisation personalisation \cite{Li2019FedMDHF} is a technique in deep learning that enhances the generalisation capabilities of a model by introducing localised perturbations to the training data. It has been shown to be effective in a variety of applications, but it is important to consider the potential risks associated with overfitting and privacy concerns.

In \cite{usynin2023sok}, the authors discuss the various aspects of memorisation in machine learning, as well as the challenges and open issues the method poses for data privacy.

Moreover, in \cite{Urvashi2019} the authors propose a method that actively enables the memorisation of unusual patterns, rather than being automatically stored in model parameters. It also shows significant improvement in performance when the prefix representations and the ML model are learned using the same training data, indicating that the prediction problem is more complex than previously thought. 

Furthermore, in \cite{Hartley2023} the authors suggest techniques to determine if a model memorises a specific (known) characteristic or not. This approach can be implemented by an outside party since it doesn't need access to the training set. The study also highlights that while memorization can affect model robustness, it can also jeopardise patient privacy when they allow their data to be used for model training.

Recent research in \cite{lesci2024causal} based on the difference-in-differences design from econometrics suggests a novel and effective approach to assess memorisation. With the use of this technique, we may define a model's memorisation profile, and its memorisation tendencies throughout training by focusing just on a limited number of training instances. It has been demonstrated that memorization in larger models is predictable from smaller ones because it is (i) stronger and more persistent in larger models, (ii) dependent on data order and learning rate, and (iii) exhibits consistent patterns across model sizes.

\subsection{Knowledge Distillation}

Generally speaking, a large difference in model size between the student and instructor networks in (KD) can lead to subpar results.
An enhanced KD framework \cite{Mirzadeh2020} was proposed, which incorporates a teacher assistant and a multi-step process. Additionally, the integration of multi-teacher KD technology with dual-stage progressive KD has been suggested \cite{Leiqi2021} to improve the performance of KD under limited data conditions. This approach takes advantage of the benefits provided by multi-teacher KD.

There have also been attempts to apply self-learning to a model via KD \cite{Yunnan2021}. The aforementioned methodology employs a teacher-student paradigm with identical network structures to derive a distilled student model. This distilled model is then leveraged as a teacher to facilitate the training of a new student model, and this cycle is iteratively repeated to gradually enhance model performance. In an attempt to avoid using exceptionally large models in Neural Machine Translation (NMT) tasks, the paper at \cite{kim-rush-2016-sequence} utilized KD, introducing two new variations of the technique in the process.

Additional variations include Relational Knowledge Distillation (RKD) \cite{park2019relational}, which transfers mutual relations between data examples. Experiments results show that via RKD, student models can often outperform the teacher. Another technique is knows as Similarity-Preserving Knowledge Distillation \cite{tung2019similaritypreserving} and it enables the training of a student network by ensuring that input pairs that generate comparable, or distinct, activations in the teacher network yield similar, or dissimilar, activations in the student network.

While exploring the field of Logit Distillation, researchers proposed a reformulation of the conventional KD loss \cite{zhao2022decoupled}, splitting it into two components referred to as Target Class Knowledge Distillation (TCKD) and Non-Target Class Knowledge Distillation (NCKD). Also a separate technique dubbed Virtual Knowledge Distillation (VKD)\cite{ Sihwan2022} leverages a softened distribution generated by a virtual knowledge generator that is conditioned on the class label, in an attempt to improve the student’s performance.

\subsection{Personalisation with Federated Learning}
AI model personalization \cite{Kairouz2021} involves adapting an AI model to a specific user or group of users. The primary goal of AI model personalization is to improve the accuracy and relevance of AI models for users. Personalization is achieved by considering the user's historical data, preferences, and behaviour patterns. AI models are designed to learn from data, and personalization involves providing the AI model with personalized data that is relevant to the user.

AI model personalization is important for several reasons. Firstly, personalization improves the accuracy of AI models. When an AI model is personalized, it is more likely to provide accurate predictions or recommendations based on the user's preferences and usage patterns. Secondly, personalization enhances the user experience. Personalized AI models are more engaging and provide users with a sense of control over the content they receive. Lastly, personalization can lead to increased revenue for businesses. Personalized AI models can help businesses to improve customer satisfaction, retention, and loyalty.

The combination of active learning and federated learning has been explored in the past. In \cite{Presotto2022} a hybrid method for Human Activity Recognition (HAR) is proposed, which relies on federated learning for collaborative model training privacy enhancement, and active learning to semi-automatically annotate the gathered data. The suggested enhanced active learning approach depends on choosing unlabeled data samples with relatively low classification confidence. VAR-UNCERTAINTY is an active learning technique that compares the prediction confidence to a dynamically adjustable threshold. In case the predicted probability value of the most likely activity is found to be below the threshold, then the user is queried to obtain the ground truth of their activity. In this work, personalization is also implemented to fine-tune the model to each user, through transfer learning strategies.

In \cite{Kelli2021} the personalization of models generated through federated learning techniques is explored for the creation of a network flow-based Intrusion Detection System (IDS) to be applied on Distributed Network Protocol 3 (DNP3)-based Supervisory Control and Data Acquisition (SCADA) systems. Initially, a global model is created in collaborative manner by the participating nodes through federated learning. However, the global model, although able to generalize, it is not adapted to the specific needs and network traffic characteristics of each participant. To that end, active learning is applied as a personalization solution, in order to customize the global model for each user in separate, after the federated training process is concluded. In this active learning scenario, the global model is trained to a small set of fully labelled samples before being introduced to a pool of unlabelled data points. The querying strategy used, namely uncertainty sampling, aims in the selection of instances for which the calculated classification uncertainty is the highest, in order to choose valuable and informative input for the model. After the most informative sample is selected, it gets labelled, and it is used to personalize the model.

Notably, a great deal of work has been done to find the best practices for collaborative and distributed machine learning in order to train federated global models in a safe, private, and efficient manner. A highly critical aspect for consideration whilst training classification and regression models for application on devices distributed across the network, is the difference in data attributes. Specifically, although data may be represented in a similar format for all devices, data values and dataset sizes may differ. Furthermore, the data amount on the nodes may be different, because some nodes produce a lot of data for model training, while other nodes produce less \cite{HUANG2022170}. In a classification scenario, this unbalance can also be described as the difference of the amount of a specific data class in each participant. This effect may be encountered due to reasons such as differences in network traffic and sensor measurements, amongst others.

Federated learning solutions are able to generate models based on the collaboration of the federated learning session’s participants, by fusing the local models trained by each node into a single, global model. However, due to the aforementioned unbalance of the data in each node, the global model is not able to perform as accurately as possible \cite{Wenjie2022}. The improvement of global models is necessary, especially in cases where the models generated through federated learning are utilized in critical sectors where high accuracy is of essence. Therefore, personalization methods should be applied after federated training, in order to ensure the betterment of global models through the customization of the federated output to each node’s needs. Notably, personalization solutions should avoid utilizing as many training data samples as a federated learning round would require securing faster training, while personalized models should be able to perform better than their federated counterpart.

One approach to personalize Federated Learning (FL) is to first train a global model on a central server using data from multiple clients, followed by fine-tuning the model’s parameters at each client using stochastic gradient descent (SGD) for a few epochs. This technique, also known as” global model fine-tuning,” allows the global model to be tailored to each client’s specific data, while still benefiting from the shared knowledge of the global model. By transmitting only model parameters rather than the entire dataset, this approach reduces the amount of data sent to the central server, preserves privacy, and enables personalized model training, potentially leading to improved accuracy \cite{jiang2020improving}, \cite{Yu2020SalvagingFL}.

Deep Neural Networks are used by Marfoq et al. \cite{Marfoq2021} to extract high-quality vectorial representations from non-tabular input like images and text. They present a mechanism for customisation via local memorization. They also demonstrate that by allowing local memorization at each Federated Learning (FL) client, it becomes possible to capture the local distribution shift of the client concerning the global distribution. In other words, our study shows that enabling the FL client to memorize its local data helps in identifying any differences between the local data distribution and the global data distribution.

In order to face the challenge that is lifelong sequential modelling and the rapid changes of user behaviour on social platforms, Ren et al. \cite{Ren2019} present the Hierarchical Periodic Memory Network, which is designed to make each user's experience memorising sequential patterns unique. This network addresses the challenge of modelling sequential data over extended periods, while also accounting for individual differences in users’ sequential patterns.

Hsieh et al. \cite{Hsieh2021} propose FL-HDC, an FL technique that introduces the bipolarizing of model parameters, which involves representing each parameter using only two bits, significantly lowering the quantity of information that must be shared between the client and the central server. To avoid loss in model accuracy, FL-HDC also includes a retraining mechanism that makes use of adaptive learning rates to make up for the accuracy loss brought on by bipolarization.

Last but not least, Lee et al \cite{Baek2020} identify a major challenge associated with Deep Learning (DL) algorithms, which is the need for high computational power and memory resources. Their proposed solution includes a technique that involves local retraining of object detectors using a new local database.

In the case of Knowledge Distillation, the authors in \cite{mora2022knowledge} present a comprehensive overview of KD-based algorithms designed to address particular FL challenges. In addition, in order to address not identically and independently distributed (non-IID) challenges, a KD-based FL framework in edge-AI called FedLCA was presented in \cite{Chaoning2023}. Both a global knowledge aggregation strategy and a local knowledge calculation strategy were put forth. Additionally, a regularisation technique based on global knowledge was offered to direct local training. Experiments have also shown us that performance can be enhanced by exchanging knowledge via the second-tolast layer of the model.

Furthermore, in \cite{Yongheng2023} a prototype-based knowledge distillation framework for FL is suggested by the authors. FedPKD allows for the collaborative learning of diverse clients and the server with varying model architectures and resource capabilities modifications by integrating prototype learning and knowledge distillation with FL. FedPKD specifically offers to transfer the dual knowledge of clients—that is, the logits and prototypes from the model output to the server—as well as a prototype-based ensemble distillation mechanism to combine the logits and prototypes from clients. This aggregated data can then be utilised to train the server model using an unlabeled public dataset. Furthermore, in order to enhance learning efficiency and minimise communication overhead, we provide a data filter mechanism based on a prototype that eliminates low-quality knowledge samples.

Moreover, through the integration of contrastive learning, FL, and rehearsal-based information distillation techniques, the work in \cite{Yongheng2023} established a comprehensive approach to minimise catastrophic forgetting and maximise knowledge retention in computer vision during incremental learning. In situations where FL has not been thoroughly researched, it offers a complete solution for ongoing learning, making it easier to learn and maintain transferable representations.

Last but not least, anew method for personalised training of local and global models in a variety of heterogeneous data environments, called "Two-fold Knowledge Distillation for non-IID Federated Learning" (FedTweet) is proposed in \cite{WANG2024109067}. In particular, to guarantee diversity in global pseudo-data, the server utilises dynamic aggregation weights for local generators based on model similarity and uses global pseudo-data for knowledge distillation to refine the first aggregated model. Clients perform adversarial training between the local model and local generator, freezing the received global model as a teacher model in the process, maintaining the personalised data in the local updates while modifying their instructions. FedTweet facilitates the exchange of teacher models across global and local models, guaranteeing mutual personalisation and generalisation.


\section{METHODOLOGY}
\label{Methodology}
\begin{figure} [h]
  \centering
  \includegraphics[width=0.4\textwidth]{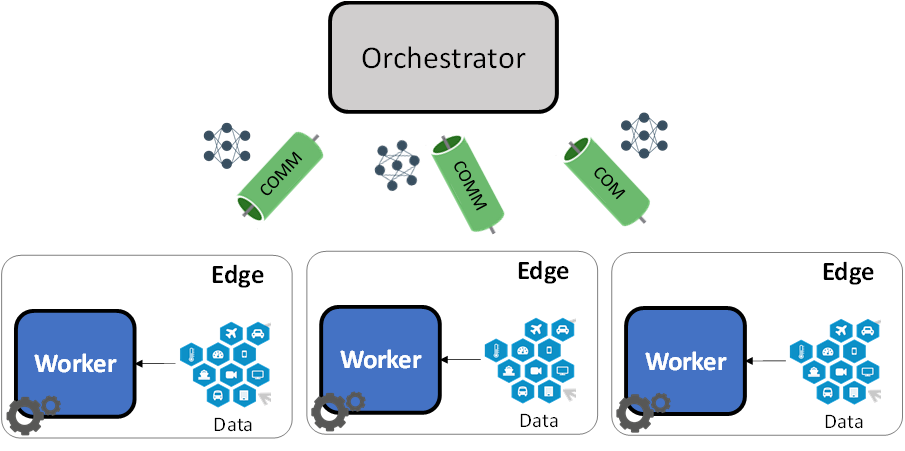}
  \caption{Centralised Federated Learning}
  \label{FIG:fig_CFLS}
\end{figure}

The core Federated Learning strategy proposed in this work is depicted in Figure \ref{FIG:fig_CFLS}. The local models are trained at the edge utilising remote devices' local data. The proposed Federated Learning approach keeps data on the edge rather than sending them to a central server in a local intranet or cloud data centre. 

A central server at a central point in the infrastructure or in the cloud orchestrates the distributed training of AI models and their fusion into one holistic global model that contains mutual knowledge from edge device training. This training scheme can be used with a very large corpus of devices, and the distributed models can be expanded horizontally (cross-device and cross-silo) and vertically (multiple security and aggregation layers), providing interoperability to a wide variety of heterogeneous environments. 

This technique uses most of the available mechanisms to secure and protect local data and its owner. This technique integrates crucial orchestration algorithms for model optimisation, resource allocation, and energy saving as the complete process is coordinated by a single point.

\subsection{Federated Learning Architecture}
\label{Federated Learning Architecture}
Contemporary computer systems are currently switching from Cloud-only implementations to edge solutions, in order to cater to the needs of the end users and offer faster services. Machine learning model training collaborative procedures in these solutions would rely on localised approach, where data would be sent by each party to a server responsible for training the aforementioned model. The introduction of the concept of edge noted that multiple distributed participants are involved while the confidentiality, integrity and availability of data exchanged was at risk. This highlighted the urgent need for a more secure and private approach to traditional model training. 

Federated learning is a distributed and collaborative model training approach with multiple participants, where instead of relying on sending data to a central entity to compute a model, it is trained locally in each party and then the weights are sent to a server for fusion and global model creation. This approach encapsulates all of the comunication, orchestration, distribution, training, and fusion of AI models from the corpus of edge devices. 
The models are trained on the collected data at each node, and the trained model weights are then sent to a global server for aggregation. The aggregation is the process of collecting and merging all of the subsidiary AI models from the edge devices into one global model, under a specific strategy and aggregation algorithm. The most commonly used aggregation algorithm is Federated Averaging \cite{McMahan2017} which undertakes the weighted averaging of the subsidiary models into the global model. Other such algorithms exist that depend on the nature of the problem and data. After this process the resulting global model is distributed back to teh edge devices for further optimisation or deployment. We can formulate the federated learning process as follows.

Initially, the global parameter server shares a global model \(w_{Global}^{0}\) along with a set of instructions on how to tain the model locally on the edge devices. These devices compose a  federated population 
\(P_{f} \in [1,N]\) where \(N \in \mathbb{N}^{*}\). Each edge device/node holds a set of local data \(D_{i \in N}\) which train the initial local model \(w_{l}^{i}\). The local models are optimised on the on-device data \(D_{i}\), and subsequently, the local model weights \(w_{Global}^{i}\) are send to the global parameter server. These weights are then aggregated using Federated Averaging (\ref{FederatedAveraging}) or a similar algorithm, resulting in a new and updated global model \(w_{Global}^{i}\) \cite{Lim2020}, which incorporates the newly acquired knowledge. Equation \ref{FederatedAveraging} summarises the process.

\begin{equation}
w_{G}^{k} = \frac{1}{\sum _{i\in N} D_{i}} \sum_{i=1}^{N}D_{i}w_{i}^{k}
\label{FederatedAveraging}
\end{equation}

Here \(w_{G}^{k}\) is the global model on the\(k_{th}\) training iteration and \(w_{i}^{k}\) is the remote \(i_{th}\) model at that iteration.

Federated learning addresses the security concerns of edge computing, however, the global model produced at the end of the federated session, although it is able to produce generalized results, is not tailored to each participant’s needs. As such, personalization of global models after federation in each node, is essential to help the model produce custom and personalized results in each node.

\begin{figure} [h]
  \centering
  \includegraphics[width=0.5\textwidth]{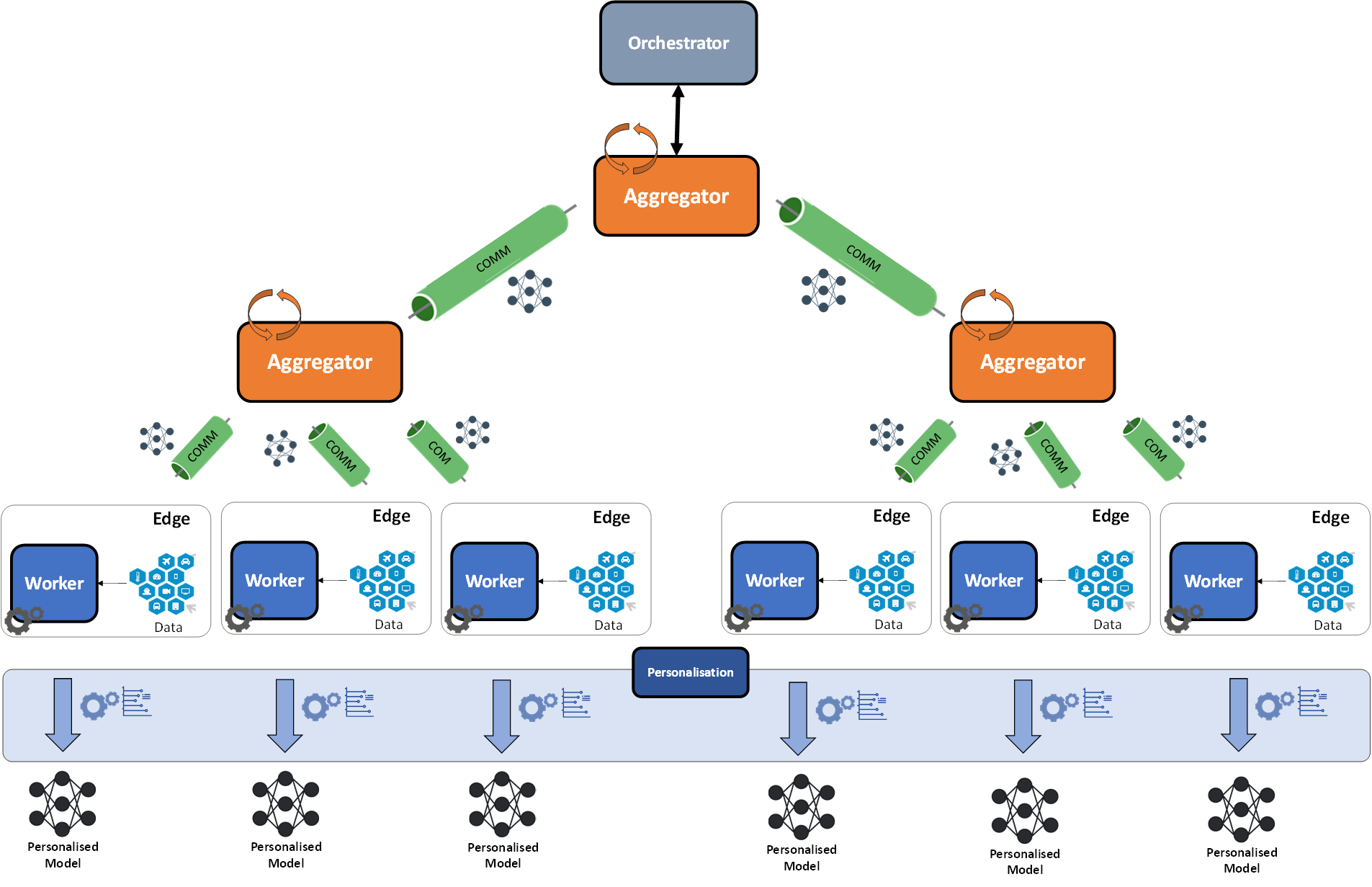}
  \caption{Centralised Federated Learning Personalisation}
  \label{FIG:fig_Centralised_Federated Learning_Personalisation}
\end{figure}

\begin{figure} [h]
  \centering
  \includegraphics[width=0.4\textwidth]{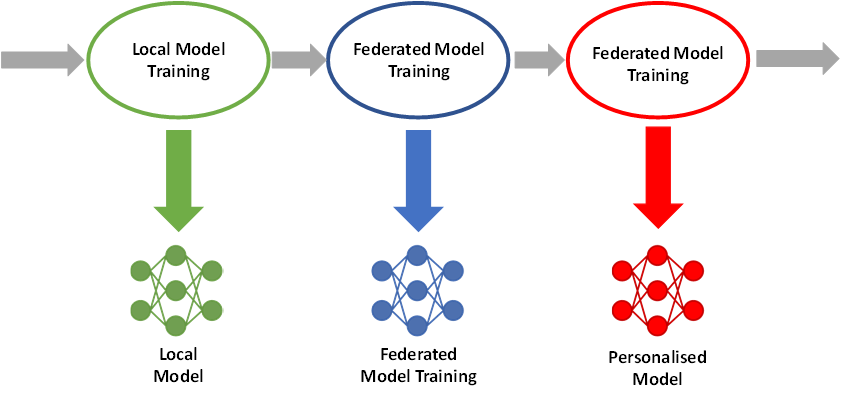}\\
  \caption{Pipeline Stages}
  \label{FIG:fig_Pipeline_Stages}
\end{figure}

The most common way to apply personalization procedures occurs after the federated learning process concludes training a global model. The model Personalisation component can be seen in Figure \ref{FIG:fig_Centralised_Federated Learning_Personalisation}. As mentioned, the personalisation of the AI models, takes place after the federation thereof. The personalisation takes place on the edge node and utilised the locally produced data. The data used can either be part of the training and testing set, but also new data that are continuously streamed to the edge node from the deployed sensors and field devices. Figure \ref{FIG:fig_Pipeline_Stages} depicts the data flow of the end-to-end process of AI model optimisation proposed in this work. 

\begin{figure} [h]
  \centering
  \includegraphics[width=0.5\textwidth]{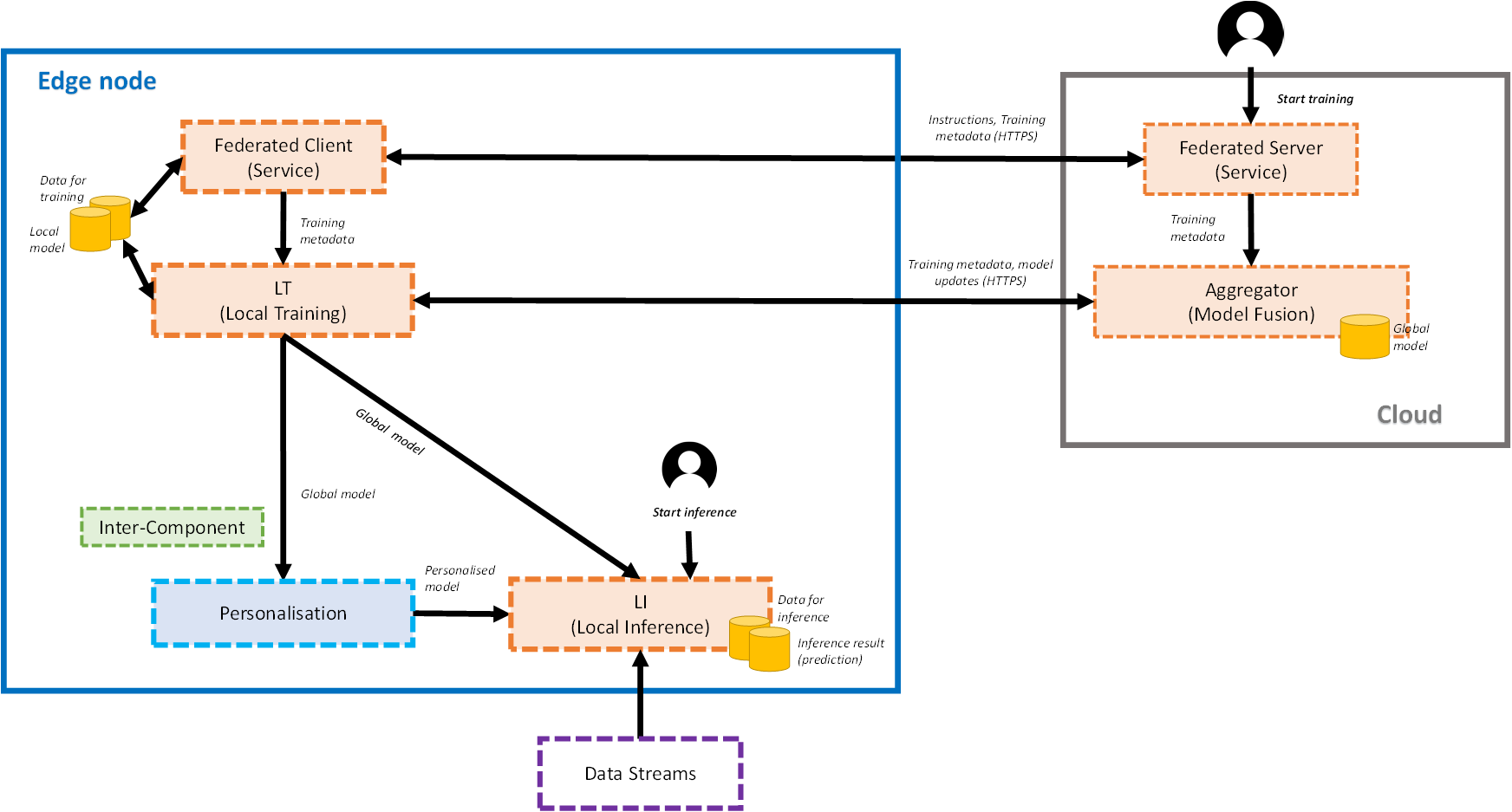}\\
  \caption{Services of Centralised FL}
  \label{FIG:fig_Services_Centralised_FML}
\end{figure}

Personalization aims to optimize and customize the global model for each participant; therefore, it is applied on each edge node of the Centralized Federated Learning approach, and by extend, it is initiated by the Federated Client service in each edge node. In the proposed approach, the Federated Client is in charge of the Local model training and the handling of the local data. Since the data never leave the Federated Client, the data processing, handling and storing is solely the responsibility of the client. The proposed centralized federated learning approach is depicted in Figure \ref{FIG:fig_Services_Centralised_FML}. 

After a federated training session is completed, the personalization of the global model is performed in all participating edge nodes. The personalization methods investigated in this work compose a process that occurs locally in each edge node and does not require any communication between the participants or orchestration by the cloud, though the personalised model can again be federated if needed. The reason for choosing this approach is to localise the adaption of the AI models to the edge device while aleviating further communication overhead that can be a possible restriction in network-constrained devices. After the model is adapted in each edge nodes’ needs, then it can be used for inferring predictions.


The overal personalisation process is divided into three sub-processes: the a) pre-processing step, the b) model training, and the model c) personalisation, depending on the utilised method. The pre-processing procedure is responsible for the transformations and adaptation of the data into features suitable for the training. The machine learning model also stems from federated learning. The model to be adapted is generated through federated learning and then passed on to the personalization component. The model personalisation is responsible for leveraging the according personalisation algorithm to further customise the AI model in the frame of the respective edge node, using the local data. In essence, the proposed Personalisation component selects the data that are valuable for the personalisation, either as a training set, or by using a sample selection process, and trains the model based on that data. We can assume the personalisation process as \ref{PersonalizedWeights},

\begin{equation}
\tilde{w}_{i}^{k} = \mathcal{P}_i(w_{i}^{k}, D_i)
\label{PersonalizedWeights}
\end{equation}

denoting the perasonalization function as $\mathcal{P}_i$ that produces a local personalized model $\tilde{w}_{i}^{k}$. Integrating the process to FL we get \ref{LocalPersonalization}, 

\begin{equation}
w_{G}^{k} = \frac{1}{\sum_{i \in N} D_{i}} \sum_{i=1}^{N} D_{i} \tilde{w}_{i}^{k}
\label{LocalPersonalization}
\end{equation}

at local iteration $k$.

\begin{figure} [h]
  \centering
  \includegraphics[width=0.4\textwidth]{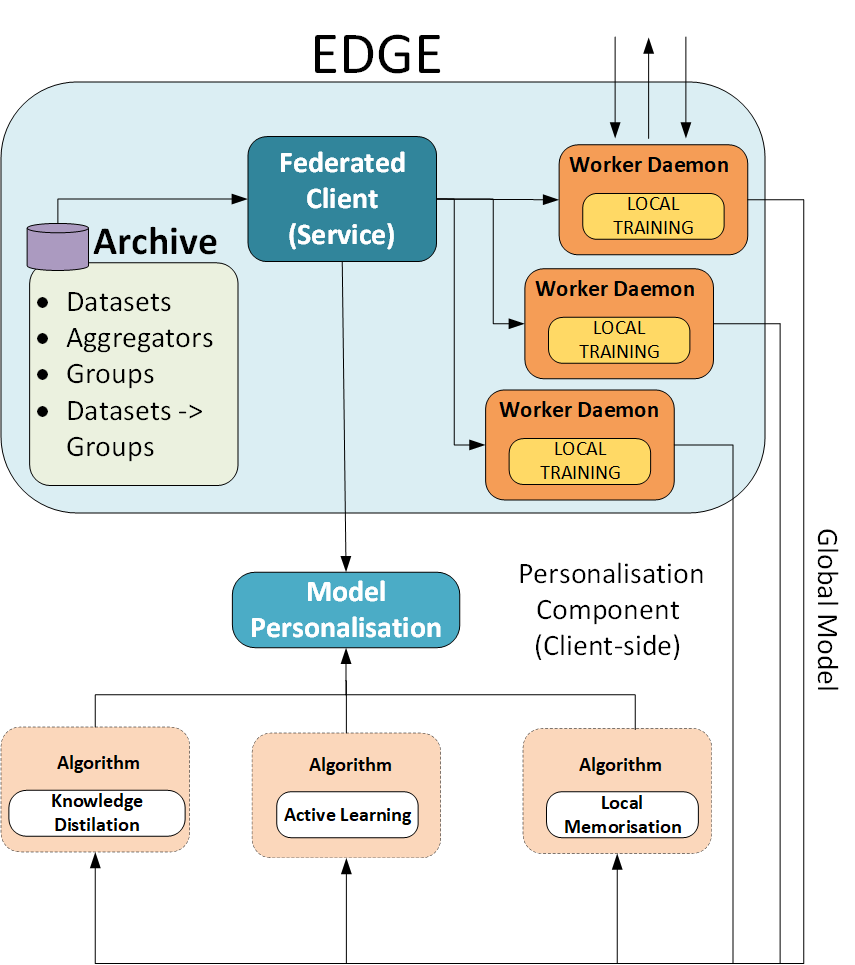}\\
  \caption{Personalisation Component in FL Architecture}
  \label{FIG:fig_PersonComp_Architecture}
\end{figure}

The interactions between the three aforementioned sub-processes of the proposed Personalisation component are depicted in Figure \ref{FIG:fig_PersonComp_Architecture}. 

For the implementation of the Personalisation component, state-of-the-art personalisation methods were leveraged, adapted and integrated into a unified component. The proposed Personalisation component encapsulates utilities to implement all of the described personalised algorithms, namely, a) Active Learning, b) Knowledge Distillation and c) Local Memorisation.

\subsection{Applying Active Learning}
\label{Applying Active Learning}
\begin{figure} [h]
  \centering
  \includegraphics[width=0.4\textwidth]{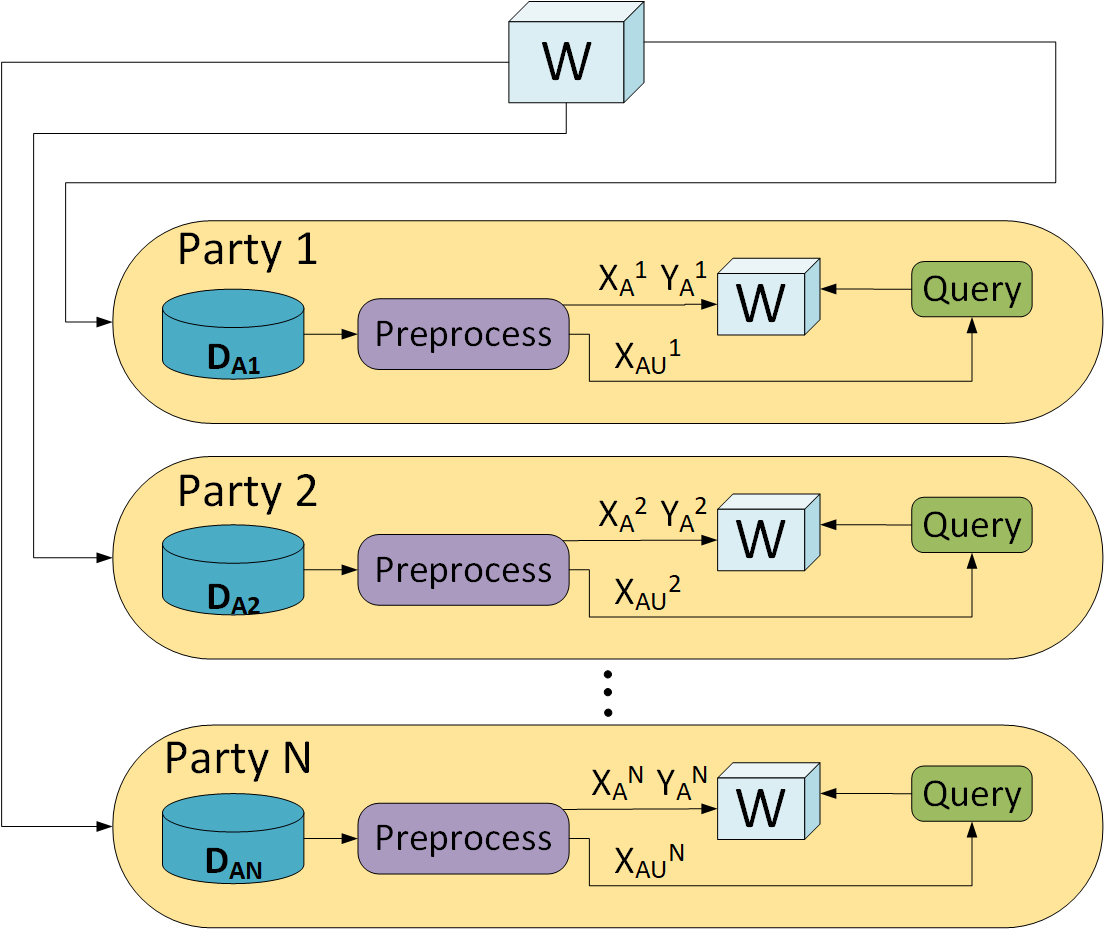}\\
  \caption{Active Learning Scheme}
  \label{FIG:fig_active_learning}
\end{figure}

\begin{figure} [h]
  \centering
  \includegraphics[width=0.3\textwidth]{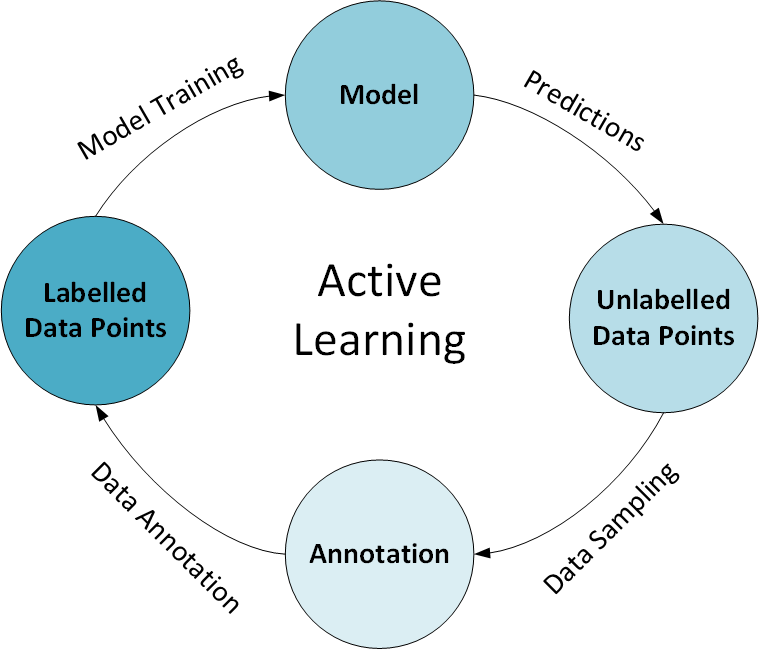}\\
  \caption{Active Learning Scheme}
  \label{FIG:fig_active_learning_process}
\end{figure}

Active Learning, as depicted in Figure \ref{FIG:fig_active_learning}, is a semi-supervised machine learning approach which allows the machine learning model, referred to as “learner” in active learning terminology, to dynamically choose samples to learn from. This means that the model itself selects training samples that it finds the most informative, in order to learn from. In the case of a classification problem, the learner selects the training sample and proceeds by querying an oracle for the provision of accurate labels. The oracle could either be a machine or a human operator. For instance, in the case of training intrusion detection systems through human supervision, the model would firstly select a training sample it deems informative, and then ask a human to label the aforementioned data sample. Next, the model gets trained by utilizing the data selected. Figure \ref{FIG:fig_active_learning_process} represents the process of active learning training.

As mentioned in the previous paragraph, this method of semi-supervised learning actively selects informative data instances to be used for training. The way that the learner assesses the training value of the data instances and chooses the most valuable data sample, is through the utilization of query strategies. One of the most-utilized technique for the selection of training points, is uncertainty-based sampling. In uncertainty-based sampling, which is a technique exploited for classification problems, the active learner selects the data instances for which it is uncertain regarding the correct label. One category of uncertainty sampling, is classification uncertainty. For example, in a binary classification problem, classification uncertainty sampling will choose the instance whose probability of being positive is nearest to 0.5. On the other hand, for multi-class classification problems, the model's confidence in prediction is used as an uncertainty measure. 
In classification uncertainty defined in the formula, the classification uncertainty \(S\) of the sample to be predicted \(x_{AUk}^p\) is calculated, with \(py_{AUk}^p\) being the most likely prediction for this instance. 
The most informative instance \(x_{AUi}^p\) is selected by picking the sample amongst the unlabeled data pool \(X_{AU}^p\) for with the classification uncertainty \(S\) is the highest. Classification margin-based sampling is another uncertainty sampling technique which calculates the difference in probability of the first and second most likely prediction. As such, the learner will select the sample with the smallest margin which would indicate the highest uncertainty. Regarding regression problems where future values are predicted, querying strategies implemented for training point selection include Gaussian solutions, where the uncertainty of the predictions is quantified, and error-based calculations where the samples that present the highest prediction error are selected.

\begin{equation}
S(x_{AUK}^p) = 1 - P(py_{AUk}^p | x_{AUk}^p)
\end{equation}


\subsection{Applying Local Memorization}
\label{Applying Local Memorization}
As explained in \cite{Tan2021TowardsPF}, local memorization personalisation is a deep learning strategy that adds localised perturbations to training data with the goal of enhancing model generalisation. In essense, local memorisation provides additional local samples to the local training of the AI model in order to enhance the global model, making it "tilt" towards the data distribution of the edge device. Local memorisation can be achieved through either providing a subset of seen $D_i^{\text{seen}}$ or unseen $D_i^{\text{unseen}}$ data by the federated training process or by selectively choosing a subset of local data $D_i^{\text{local}}$ that are representative of the local data distribution. all of the data belong to the clinet data $[D_i^{\text{seen}}, D_i^{\text{unseen}}, D_i^{\text{local}}] \in D_i$. We can further describe the relationship of these subsets in the context of the personalisation of the AI model by including them in the overall process. We can add a weight (proportions) of each of this subsets in relation to the personalised model $\tilde{w}_{i}^{k}$, as follows:

\begin{equation}
\tilde{w}_{i}^{k} = \alpha w_i^{k} (D_i^{\text{seen}}) + \beta w_i^{k} (D_i^{\text{unseen}}) + \gamma w_i^{k} (D_i^{\text{local}})
\label{LocalMemorization}
\end{equation}

Where $w_i^{k} (D_i^{\text{seen}})$ denotes the global model weights trained on the seen data subset, $w_i^{k} (D_i^{\text{unseen}})$ trained on the unseen data subset and $w_i^{k} (D_i^{\text{local}})$ to the on the representative local data subset, respectively. Additionally, we include that $\alpha, \beta, \gamma \in [0,1]$, while $\alpha+\beta+\gamma=1$, as we can use any needed proportion of these subsets. 

A very obvious advantage of this method is that it does not require additional computation for the edge device to compute the optimal training vectors, like active learning, it just requires minimal training from the client's side to provide the aforementioned "tilt" to the model. Though useful in many contexts, it is important to evaluate possible hazards such overfitting, that can be tackled by selective finetuning.

\subsection{Applying Knowledge Distillation}
\label{Applying Knowledge Distillation}
\begin{figure} [h]
  \centering
  \includegraphics[width=0.47\textwidth]{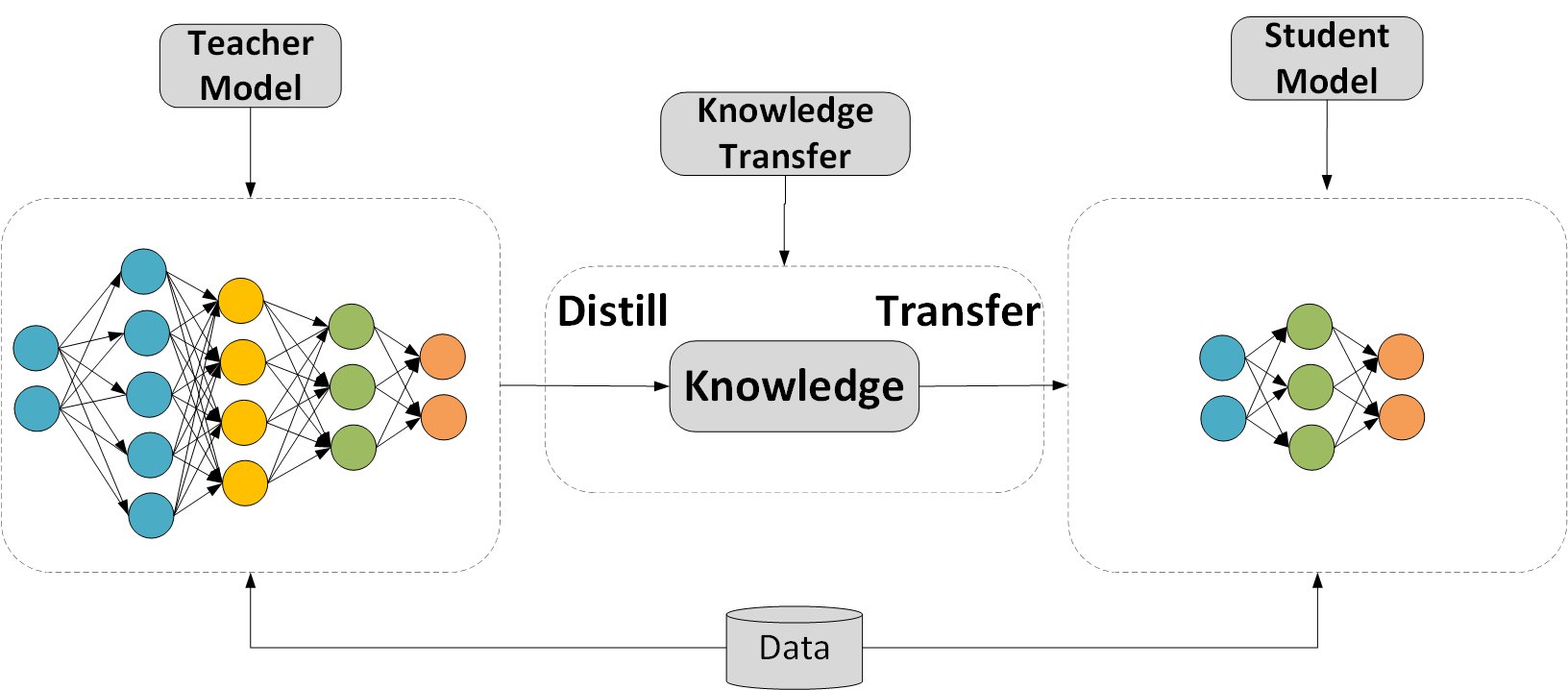}\\
  \caption{Knowledge Distillation Scheme}
  \label{FIG:fig_KDS}
\end{figure}

\begin{figure} [h]
  \centering
  \includegraphics[width=0.4\textwidth]{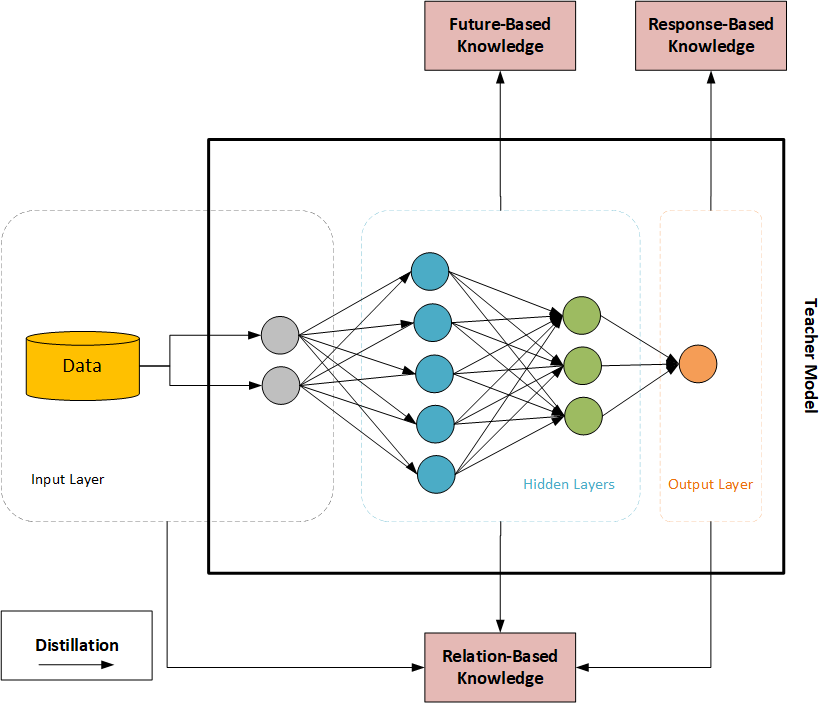}\\
  \caption{Knowledge Distillation Process}
  \label{FIG:fig_KDP}
\end{figure}

According to \cite{zhao2022decoupled}, Knowledge Distillation (KD) is a model personalisation method used to move knowledge from a sophisticated teacher model to a more straightforward student model (Figures \ref{FIG:fig_KDS} and \ref{FIG:fig_KDP}). 

Past research results \cite{zhao2022decoupled} have shown that using large models as teachers often leads to suboptimal results. A proposed solution to this problem is the early termination of the teacher’s training. According to the same source \cite{zhao2022decoupled}, the process of KD is as follows. Let us consider a collection of cases in the form of $(x, y)$, where $y$ belongs to a set of possible classes $V$, to train a multi-class classifier. The objective of training a model is to minimize loss, the difference between predictions and real values, for each instance of the training data.

The following is the KD process. Consider the scenario where we are training a multi-class classifier on a dataset of samples represented as \((x,y)\) with \(V\) as possible classes. The objective of training a model is to minimize loss, the difference between predictions and real values, for each instance of the training data.

\begin{equation}
L(\theta) = - \sum^{|V|}_{k=1} I ( y = k ) log p( y = k |x;\theta)
\end{equation}

Here, the symbol \(I\) represents the indicator function, and \(p\) denotes the distribution from our model that is parameterized by \(\theta\). The goal is to minimize the cross-entropy between the distribution of the model distribution \(p(y /  x;\theta)\) and the degenerate data.

Assuming access to a learned teacher distribution \(q(y /  x;\theta t)\), which may have been trained on the same data set, the approach involves minimizing the cross-entropy with the teacher's probability distribution instead of with the observed data.

\begin{equation}
LKD(\theta ; \theta t) = - \sum^{|V|}_{k=1} q ( y = k | x ;  \theta t ) log gp( y = k |x;\theta)
\end{equation}

In which the parameter \(\theta t\) is used to define the teacher distribution and is kept constant. The cross-entropy setup remains the same, but the target distribution is no longer a sparse distribution.

Given the absence of a direct term for the training data in the new objective, it is widely accepted to apply an interpolation technique that blends between the two losses.

\begin{equation}
L(\theta ; \theta t) = - (1-a) L (\theta) + aLKD( \theta ;\theta t)
\end{equation}

In the above formula, \(a\) represents a mixture coefficient that combines the one-hot distribution and the teacher distribution.

\subsection{ML algorithms for Personalisation Refinement}
\label{ML algorithms for Personalisation Refinement}
The main challenge presented is data regression and/or future value forecasting. To that end, four main regression methods, namely, i) Linear Regression, ii) Deep Neural Network (DNN), iii) Long-Short Term Memory (LSTM) network and iv) Transformer network, were implemented and tested against both the Federated Learning and Personalisation components. Out of the four selected methods, the first method was utilised as a baseline, due to its statical linearity and the fact that it is commonly used for solving value forecasting problems.

\subsubsection{Linear Regression}
\label{Linear Regression}
One of the most basic types of models, and the simplest in the present collection. It consists of only an input and an output layer. While it was possible to enhance the model with hidden layers, it remained simple to be used as a comparison point. The model was utilized for a regression issue, but it’s also useful for classification problems with some slight changes.

\begin{figure} [h]
  \centering
  \includegraphics[width=0.1\textwidth]{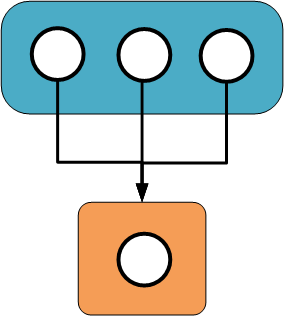}\\
  \caption{Linear Regressor}
  \label{FIG:fig_Linear_Regressor}
\end{figure}

This model exclusively uses the linear activation function for its output layer, which means the model can’t learn complicated relationships between its input and output. The aforementioned function is showcased in Figure \ref{FIG:fig_Linear_Regressor} and in the following equation:

\begin{equation}
f (x) = x
\end{equation}

\subsubsection{Deep Neural Network (DNN)}
\label{Deep Neural Network (DNN)}
It’s a common occurrence for ANN models to contain hidden layers, with the intent of boosting the overall model performance. The inclusion of hidden layers in a neural network enables the retention of information that governs the input's relevance to the output. Networks consisting of multiple hidden layers, typically two or more, are classified as Deep Neural Networks (DNNs). Each layer in a DNN is fully connected, meaning that every neuron in the layer below is connected to every other neuron in the layer above.

\begin{figure} [h]
  \centering
  \includegraphics[width=0.2\textwidth]{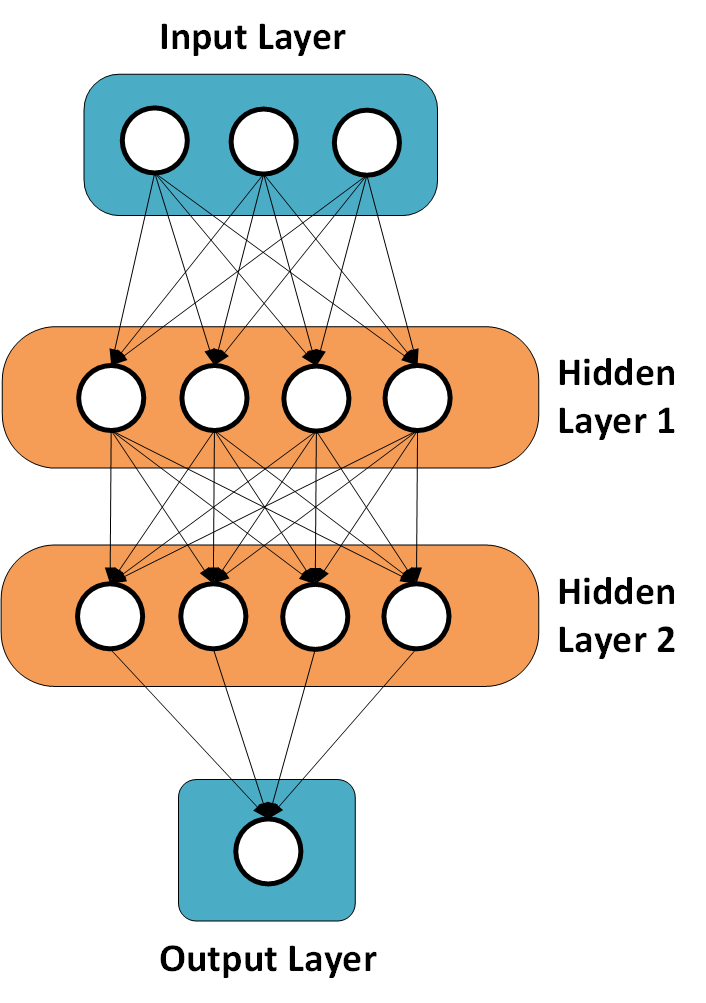}\\
  \caption{Simple Deep Neural Network}
  \label{FIG:fig_Simple_DNN}
\end{figure}

For this model and the subsequent ones, the ReLU activation function is employed instead of Linear. This function introduces non-linearity into the relationship from each layer's input to its output, which enhances the model's ability to address more intricate problems. The ReLU activation function is displayed in Figure \ref{FIG:fig_Simple_DNN} and at the following equation:

\begin{equation}
f (x) = max(0,x)
\end{equation}

DNNs have demonstrated significant success, leading to the emergence of various subcategories of deep learning models, such as Convolutional Neural Networks (CNNs), which excel in image recognition, and Recurrent Neural Networks (RNNs), which have gained popularity for their superior capacity in handling sequential and time series data, surpassing traditional network models. For our purposes, the DNN employed retains a simplistic structure with a limited number of hidden layers, intended solely for performance comparison with LSTMs.

\subsubsection{Long-Short Term Memory (LSTM)}
\label{Long-Short Term Memory (LSTM)}
\begin{figure} [h]
  \centering
  \includegraphics[width=0.5\textwidth]{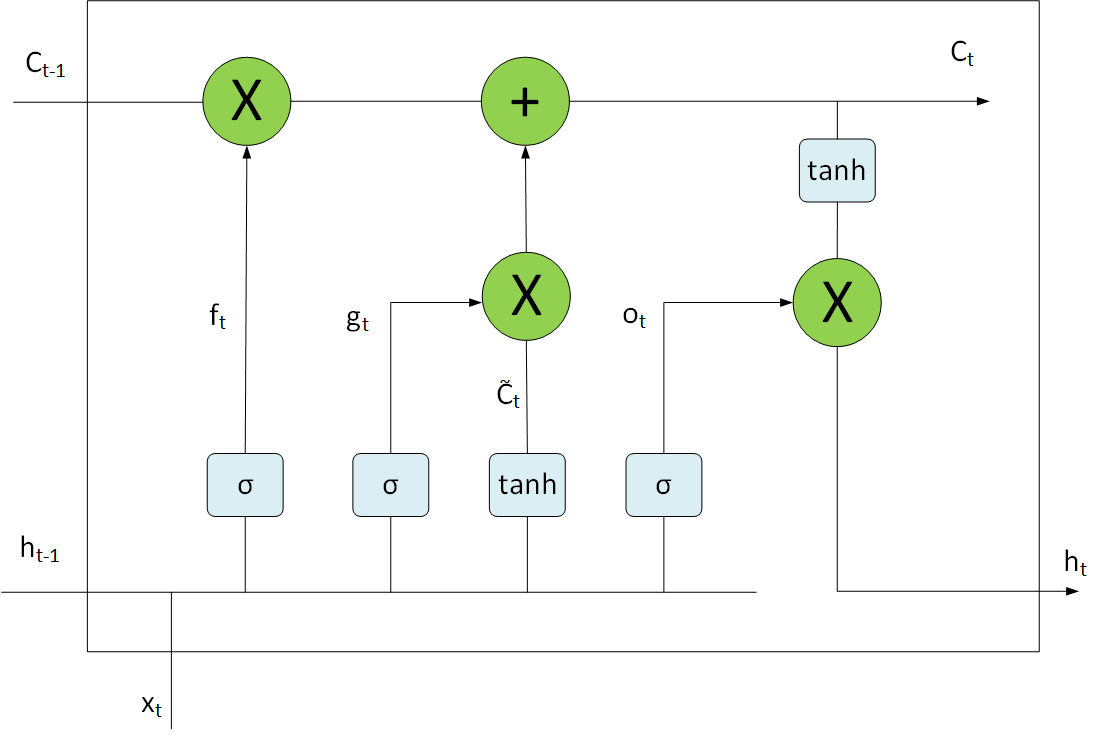}\\
  \caption{Long-Short Term Memory Network}
  \label{FIG:fig_Long_Short_Term_Memory_Network}
\end{figure}

The success of RNNs is attributed to their ability to retain and apply context throughout the prediction process. However, the memory of RNNs is restricted to short-term storage, leading to underperformance when the context exceeds the limit of the model's memory. As the memory reaches its limit, the oldest retained information is replaced with newly received data.
Long Short-Term Memory (LSTM) models are a widely adopted variation of conventional RNNs, specifically designed to address the memory limitations of the latter and facilitate more efficient context handling. LSTMs work by storing important data sequences in their short-term memory while discarding unneeded information.
The process undertaken by LSTMs can be viewed in Figure \ref{FIG:fig_Long_Short_Term_Memory_Network}. The top line of represents the short-term memory, referred to as the Cell State \((C_{t-1}, C_{t})\), where crucial information is retained. The bottom line of an LSTM consists of the input \((x_{t})\)  and the hidden state \((h_{t-1}, h_{t})\), which is a shared feature across various types of models. An updated hidden state and cell state are the output of each LSTM cell. In a single LSTM cell, the input, previous hidden state, and cell state are processed through a series of gates. These are:
• the forget gate \((f_{t})\),
• the input gate \((g_{t})\)  and
• the output gate \((o_{t})\).

Information is either stored or destroyed based on the forget gate. In addition to the current state, this also contains the cell state and concealed state from the preceding loop. The forget gate uses the following equation:

\begin{equation}
f_{t} = \sigma (W_{f} [h_{t-1},x_{t}] + b_{f})
\end{equation}

The sigmoid function \(\sigma\) is applied to the input \((x_{t})\) and the preceding hidden state \((h_{t-1})\). The result is a vector with normalised values in the [0, 1] range that is the same size as the preceding cell state. Every element in the input and the preceding concealed state is given a value between $[0,1]$ using the sigmoid function. A number of $0$ indicates total forgetfulness of past knowledge, whereas a value of $1$ indicates total retention of past knowledge. Finally, \(b\) and \(W\) represent this gate's bias and cyclic weights, respectively.
The input gate determines how much is needed to complete the task by evaluating the current input value. 

 This process involves two stages, the first being shown in the following Equation, which employs a sigmoid layer to determine which values to retain, either $0$ or $1$:

\begin{equation}
g_{t} = \sigma (W_{g} |g_{t-1}, x_{t}| + b_{g})
\end{equation}

Then, the second step, which is viewable in the following Equation , involves the use of a tanh layer, which assigns weight to all the retained data, giving them a significance value:

\begin{equation}
\tilde{C}_{t} = tanh (W_{s} [h_{t-1}, x_{t}] + b_{s})
\end{equation}

Subsequently, all the information deemed valuable by the input gate is added to the cell state \(C_{t}\), which is then utilized as the updated cell state from that point onwards. The cell state is updated through the following Equation:

\begin{equation}
C_{t} = f_{t} C_{t-1} + g_{t} \tilde{C}_{t}
\end{equation}

The output gate, predictably, determines which values to output. This process is also split into two steps, with the first one, shown in the following Equation, involving the execution of a sigmoid layer to determine which data can pass through:

\begin{equation}
o_{t} = \sigma (W_{o} [h_{t-1}, x_{t}] + b_{o})
\end{equation}

The updated cell state is then multiplied by the sigmoid output after being resampled to $[-1, 1]$ using a tanh layer. The new hidden state is where this procedure is realized, via the following Equation:

\begin{equation}
h_{t} = tanh(C_{t}) o_{t}
\end{equation}

\subsubsection{Transformer}
\label{Transformer}
The term Transformer refers to a type of Deep Learning which employs an encoder-decoder arrangement for their design. Transformers are built to simultaneously comprehend the relationships between each component of a sequence. Transformers may be more successful in capturing enduring relationships and connections across different sequence parts, maintaining context in a DL work with little to no constraints.

The encoder's job of converting input data into a fixed-length vector frequently places restrictions on encoder-decoder designs. This constraint may lead to the decoder processing the data insufficiently. On the other hand, Transformers utilize an attention mechanism \cite{Bahdanau2014NeuralMT} that enables the network to concentrate on specific sections of the input stream, thereby improving the model's efficacy in generating an output. The encoder and decoder work together in the Transformer design to convert the input sequence into a vector that contains all of the contextual information for the entire sequence. The decoder, on the other hand, is in charge of decoding this context and producing useful output.

The arrangement of components in a deep learning task determines the sequence. While models that process data sequentially don't encounter any issues, Transformers must assign a relative position to each component. This requirement is addressed through a technique called positional encoding \cite{Vaswani2017AttentionIA}. This procedure makes strategic use of the sine and cosine functions. For every even index, the sine function produces a vector, and for every odd index, the cosine function produces a vector. In this case, \(pos\) indicates the element's position inside the sequence, \(i\) the dimension's index in the embedding vector, and \(d_{model}\) the dimensionality of the embedding. The matching sections of the input sequence are then supplemented with these generated vectors.

\begin{equation}
PE_{(pos, 2i)} = sin(pos/10000^{(2i/d_{model})})
\end{equation}
\begin{equation}
PE_{(pos, 2i+1)} = cos(pos/10000^{(2i/d_{model})})
\end{equation}

\begin{figure*} [h]
  \centering
  \includegraphics[width=0.9\textwidth]{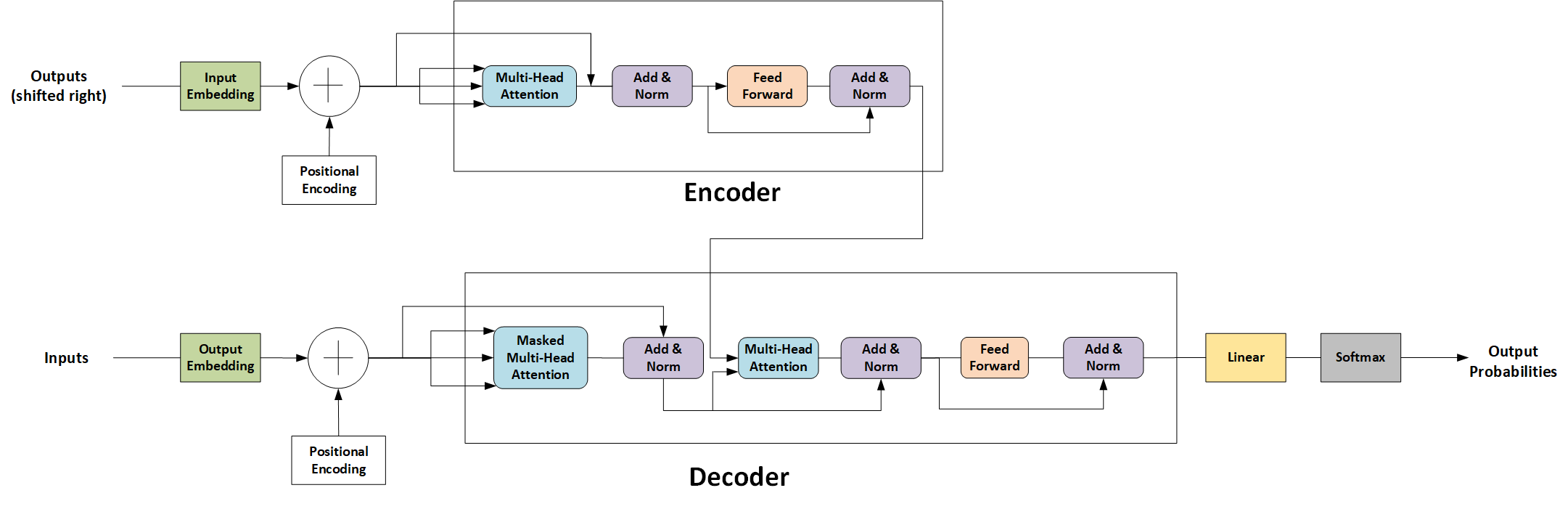}\\
  \caption{Transformer Memory Scheme}
  \label{FIG:fig_Transformer}
\end{figure*}

The Transformer architecture as a whole can be viewed in Figure \ref{FIG:fig_Transformer}. The multi-headed attention sub-module and a completely connected network are the two sub-modules that make up the encoder layer.
Both sub-layers are accompanied by residual connections and a normalization layer. By enabling the Encoder's self-attention mechanism, the multi-headed attention module makes it easier for each input element to connect with other elements in the sequence.

\begin{figure} [h]
  \centering
  \includegraphics[width=0.5\textwidth]{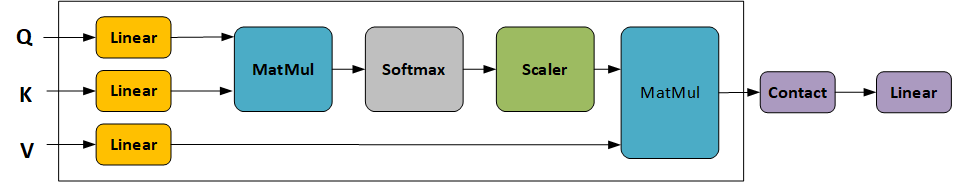}\\
  \caption{Self-Attention Process}
  \label{FIG:fig_Self_Attention_Process}
\end{figure}

The process of self-attention is shown in Figure \ref{FIG:fig_Self_Attention_Process}. The Transformer processes the information through three distinct yet interconnected layers in order to attain self-attention:
\begin{itemize}
    \item The Query \(Q\)
    \item The Key \(K\)
    \item The Values \(V\)
\end{itemize}

The Q, K, and V vectors are first transformed via separate linear layers. The attention score is then computed, which assesses each data component's importance in relation to the others. Dot product operations must be carried out between the $Q$ vector and each of the $K$ vectors in the data sequence.
The scores are divided by the square root of the dimensions of the query and key vectors in order to scale them down and improve gradient stability. The scaled scores are then subjected to the softmax function, which generates attention weights expressed by probabilities $p \in [0,1]$. The entire process of determining the attention score is denoted in the following Function:

\begin{equation}
Attention(Q,K,V) = soft max (\frac{QK^T}{\sqrt{d_{k}}})
\end{equation}

\begin{figure} [h]
  \centering
  \includegraphics[width=0.5\textwidth]{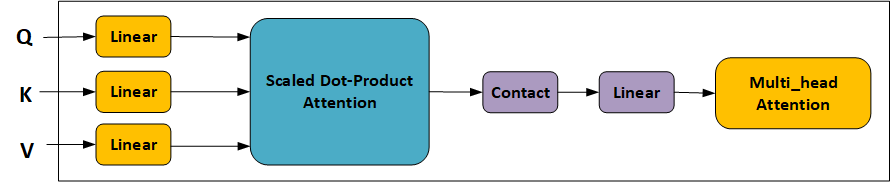}\\
  \caption{Multi-head Attention}
  \label{FIG:fig_Multi_head_Attention}
\end{figure}

The attention mechanism process can be paralleled, enabling a more advanced version of it called multi-headed attention. In essence, this implies that attention can be utilized in a collaborative manner by multiple processes (Figure \ref{FIG:fig_Multi_head_Attention}). The objective is for each head to learn unique information, thereby expanding the encoder's capabilities. To achieve this, separate the query, key, and values are split into several sub-vectors before using self-attention. Every self-attentional event is referred to as a head, and every head produces an output vector. Before being sent through the last linear layer, these output vectors are combined into a single vector.

\begin{figure} [h]
  \centering
  \includegraphics[width=0.5\textwidth]{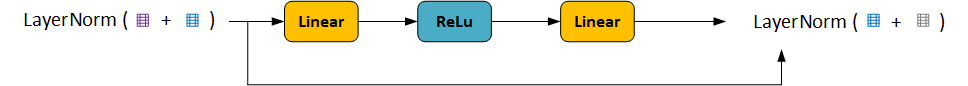}\\
  \caption{The input and output of the pointwise feed-forward layer}
  \label{FIG:fig_pointwise_feed_forward_layer}
\end{figure}

A residual connection is formed by merging the multi-headed attention output vector with the initial positional input embedding. Afterward, the output of the residual connection undergoes layer normalization and is projected through a pointwise feed-forward network (Figure \ref{FIG:fig_pointwise_feed_forward_layer}) for further processing. The aforementioned network consists of a ReLU activation function sandwiched between linear layers, its output normalized and added to the normalization of the layer.

The Decoder is made up of two multi-headed attention layers, a layer applied after each sub-layer for layer normalization, residual connections, and a pointwise feed-forward layer. The Decoder utilizes an auto-regressive approach to generate tokens sequentially while taking inputs, starting with a special start token and outputting a new token. Each multi-headed attention layer has a unique role, with the first layer computing attention scores for the input using positional embeddings. Additionally, the output of the Encoder offers crucial attention-related information to the Decoder, while the final linear layer performs as a classifier with softmax acquiring component probabilities.

A masking procedure is used in the first attention layer to avoid conditioning upcoming tokens. Using a process called look-ahead masking, this alters the attention ratings for future tokens by changing them to "-inf".

\begin{figure} [h]
  \centering
  \includegraphics[width=0.5\textwidth]{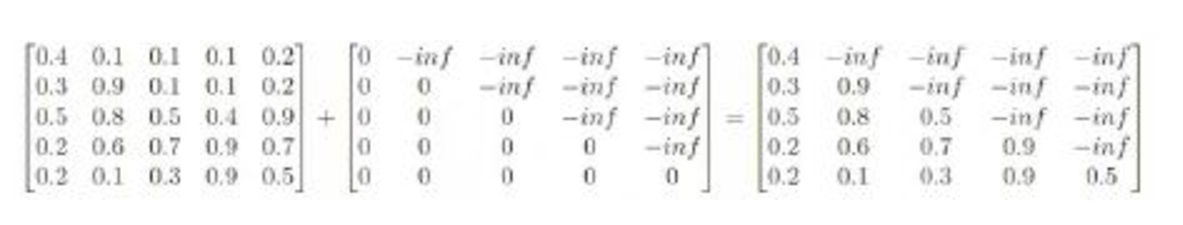}\\
  \caption{Mask Matrix}
  \label{FIG:fig_matrix}
\end{figure}

The initial matrix displays the scores that have been scaled, while the subsequent matrix depicts the application of a look-ahead mask to these scores. The resulting matrix presents the adjusted scores after the mask has been applied (Figure \ref{FIG:fig_matrix}).

The attention-giving procedure occurs during the masking process between the scaler and softmax layers. By giving future token values a zero value and hence removing their influence, the softmax function normalises the new scores produced by look-ahead masking.
The second attention layer of the Decoder focuses on the most important information by aligning the input from the Encoder with the input of the Decoder. After going through additional processing, the output of this layer is fed into a linear layer and a softmax layer, where the prediction is made using the index with the highest score.
The Decoder can be layered in layers to improve its prediction power. 
This method increases variability and enables the Decoder to take into account more data when making predictions.

\section{EVALUATION}
\label{Evaluation}


\subsection{Evaluation Data}
\label{Datasets}
In order to test and validate the technical implementation of the aforementioned components, both Federated Learning and Personalisation experiments were conducted. To ensure that the experiments were comprehensive and realistic, a variety of heterogeneous datasets were used from different domains such as healthcare, agriculture, and industry.
The utilisation of diverse datasets with varying characteristics ensured that the experiments were conducted in a realistic setting, as the data employed in the experiments was not restricted to a singular domain or application. The utilisation of a diverse range of data facilitated a more precise assessment of the efficacy of the Federated Learning and Personalisation models in terms of their performance. 

The use of diverse datasets also enabled the evaluation of the Federated Learning and Personalisation models across multiple domains. This is important as it allows for the identification of any limitations or challenges that may arise when implementing these models in different contexts. By conducting experiments on a diverse range of datasets, the results can be used to inform the development of more robust and adaptable AI models that can be applied to various use cases and domains.

\subsubsection{Smart Agriculture Data}
The Smart Agriculture Data is a collection of temporal information compiled from sensor readings taken inside stables. One stable is represented by each node. The goal is to forecast future stable conditions using the sensor data that has been provided and to ensure the well-being of the animals kept in those stables. 
\begin{itemize}
    \item Farm Animal Welfare, in Table \ref{table:Animal Welfare Features}, is a batch of artificially produced datasets based on farm animal welfare that were made to provide additional data to the model to aid in generalization. They were treated as separated dependencies from the real datasets.
    \item The Animal Feed Cultivation, in Table \ref{table:Animal Feed Features}, is provided from sensors that have been deployed in the field and are currently located in various remote areas. The conditions of various crops and plantations of decentralized agricultural infrastructures are surveyed by these sensors. The nodes offered are not uniform. The goal is to forecast future field conditions in order to assist farmers in lowering production costs and maximizing crop yield.
\end{itemize}

\begin{table}[h]
\caption{Animal Cultivation Dataset - Animal Welfare Features}
\label{table:Animal Welfare Features}
\begin{tabular}{|p{3cm}p{7cm}|}
\hline
\multicolumn{2}{|l|}{\textbf{Farm Animal Welfare}}                                  \\ \hline
\multicolumn{1}{|l|}{Features}    & DateTime, Air Humidity, Air Temp, Ch4, CO2 Avg, CO2 Max, CO2 Min, Counter, Dew Point Temp  \\ \hline
\multicolumn{1}{|l|}{Records}     & \textgreater{} 80K                          \\ \hline
\multicolumn{1}{|l|}{Nodes}       & 2       \\ \hline
\end{tabular}
\end{table}

\begin{table}[h]
\caption{Animal Cultivation Dataset - Animal Feed Features}
\label{table:Animal Feed Features}
\begin{tabular}{|p{3cm}p{7cm}|}
\hline
\multicolumn{2}{|l|}{\textbf{Animal Feed Cultivation}}                                  \\ \hline
\multicolumn{1}{|l|}{Features}    & DateTime, Air Humidity, • Air Pressure • Air Temperature • Battery • Counter • Dew Point Temp, Volumetric WC, Soil Temp  \\ \hline
\multicolumn{1}{|l|}{Records}     & \textgreater{} 3M                           \\ \hline
\multicolumn{1}{|l|}{Nodes}       & 6       \\ \hline
\end{tabular}
\end{table}

\subsubsection{Smart Home Data}
The Smart Home Data contains information from both internal and external environmental conditions within Smart Buildings.
\begin{itemize}
    \item The Electricity Smart Meter Data, in Table \ref{table:Electricity Data Features}, contains information characterizing energy loads and consumption patterns within Smart Buildings. The objective is to anticipate forthcoming energy load consumption and production through analysis of the available data. 
    \item The Smart Building Energy Management Data, in Table \ref{table:Smart Building Energy Management}, contains monitoring data regarding temperature and dimming levels.
\end{itemize}

\begin{table}[h]
\caption{Smart Home Dataset - Electricity Data Features}
\label{table:Electricity Data Features}
\begin{tabular}{|p{3cm}p{7cm}|}
\hline
\multicolumn{2}{|l|}{\textbf{Electricity Smart Meter Data}}                                      \\ \hline
\multicolumn{1}{|l|}{Features}    & eventDate, VAR_S, PF_L1, PF_L2, VA_S, PF_L3, V_L2_N, VA_L2, V_L3_N, VA_L3, V_L1_N, VA_L1, VAR_L3, VAR_L2, W_L1, VAR_L1, W_L2, W_L3, W_S, Wh_S, PF_S, Hz, A_L2, A_L3, A_L1, VArh_Ind_S, VArh_Cap_S    \\ \hline
\multicolumn{1}{|l|}{Records}     & \textgreater{}360M                             \\ \hline
\multicolumn{1}{|l|}{Nodes}       & 2       \\ \hline
\end{tabular}
\end{table}

\begin{table}[h]
\caption{Smart Home Dataset - Smart Building Energy Management Features}
\label{table:Smart Building Energy Management}
\begin{tabular}{|p{3cm}p{7cm}|}
\hline
\multicolumn{2}{|l|}{\textbf{Smart Building Energy Management Data}}                    \\ \hline
\multicolumn{1}{|l|}{Features}    & eventDate, setTemp, operationMode, userControl, fanSpeed, tempAct, status, accumulatedPower, dimming, luminance, temperature, humidity, gustWindSpeed, averageWindSpeed, airTemperature, solarRadiation, airHumidity, windDirection                   \\ \hline
\multicolumn{1}{|l|}{Records}     & \textgreater{}79K                                    \\ \hline
\multicolumn{1}{|l|}{Nodes}       & 2              \\ \hline
\end{tabular}
\end{table}

\subsubsection{Supply Chain Data}
The Daily Product Sales, in Table \ref{table:Dairy Product Sales}, has the objective to forecast upcoming product sales using current data. Seven different products’ three-year sales data were made available, and each one was regarded as an independent node. The databases provide a variety of information on daily product unit sales.
\begin{table}[h]
\caption{Supply Chain Dataset - Dairy Product Sales Features}
\label{table:Dairy Product Sales}
\begin{tabular}{|p{3cm}p{7cm}|}
\hline
\multicolumn{2}{|l|}{\textbf{Daily Product Sales}}  \\ \hline
\multicolumn{1}{|l|}{Features}    & Day, Month, Year, Daily Sales, Daily Sales (Previous Year), Daily Sales (percentage difference), Daily Sales KG, Daily Sales KG (Previous Year), Daily Sales KG (percentage difference), Daily Returns KG,  Daily Returns KG (Previous Year), Points of Distribution, Points of Distribution (Previous Year)              \\ \hline
\multicolumn{1}{|l|}{Records}     & \textgreater{}7K                                    \\ \hline
\multicolumn{1}{|l|}{Nodes}       & 7            \\ \hline
\end{tabular}
\end{table}

\subsection{Evaluation Metrics}
Using multiple criteria, the tested models' predictions were compared to the datasets' true values to select the best model. To maintain consistency and impartiality in evaluation, these criteria were employed throughout the trial cycle. Below is a list of measures that were valid throughout the trial. The models' performance was calculated using MAE and MSE. These metrics are commonly used in machine learning to assess prediction accuracy. Over all data points (n), the MAE is generated by averaging the absolute differences between the real values (xi) and the predicted values (yi). All data points' squared discrepancies between true and forecasted values are averaged to produce the MSE. Both measures quantify the dataset's projected values vs its actual values.

Root mean squared error was also employed to assess model performance. Taking the square root of the MSE gives the RMSE, which measures prediction error standard deviation. Furthermore, a variety of criteria were used to evaluate the models' accuracy and efficacy. This information may be used to choose the optimal model for the job and dataset and improve machine learning algorithms in the future.

These metrics were chosen both for their disposition in accurately quantifying the performance of forecasting AI models but also due to their wide adoption by the multitude of implementation in the respective domains. The adopted metrics are analysed below:

I. Mean Absolute Error (MAE): Is the average of all of the differences that exist between the actual and the projected values. The fact that the distinction between them may be easily understood contributed to its widespread use. We may see the mathematical formula for MAE written out in the following Equation.

\begin{equation}\label{mae-equation}
    MAE = \frac{1}{n} \sum_{i=1}^{n}|T_i - P_i|
\end{equation}

II. Mean Square Error (MSE): It works out the mean squared deviation between the values that were anticipated and those that were actually observed. When a model has no error, MSE = 0. The formula for MSE is shown here.

\begin{equation}\label{mse-equation}
    MSE = \frac{1}{n} \sum_{i=1}^{n}(T_i - P_i)^2
\end{equation}

III. Root Mean Square Error (RMSE): The value of the MSE expressed as its square root. As a result of the fact that it is measured using the same units as the answer variable, it is possible to understand it in a straightforward manner and is hence frequently seen as a more accurate evaluative tool. The following displays the formula for calculating RMSE.

\begin{equation}\label{rmse-equation}
    RMSE = \sqrt{\frac{1}{n} \sum_{i=1}^{n}(T_i - P_i)^2}
\end{equation}

It is worth noting that in the case of value regression, the metrics utilised are relative to the case and data of the respective implementations. In particular, depending on the data distribution and scale, some of the abovementioned metrics might produce invalid measurements such as infinity values. In those cases the values are normalised to [0.0] for uniformity. If the overall results of a certain metric on an experiment are perceived as invalid, the metric may be omitted.




\subsection{Experiment Results}
\label{Experiment_Results}
The experiments were performed in two phases: a) Local training, to provide a baseline of the performance of the non-Federated AI model training and b) Federated training, where the local models were federated to create a holistic and optimised global model.

A series of additional experiments were conducted, using the personalisation algorithms analysed before, a) Active Learning, b) Knowledge Distillation and c) Local Memorisation. These experiments are aimed to validate the methods and techniques employed and implemented for personalization. The set of these experiments served as a baseline to quantify the performance optimization and efficacy offered by the personalization module.

The experiments were conducted in sequence. First the model was trained locally, to establish the performance of the non-personalized AI model. Subsequently, the distributed local models were trained using the Federated Learning process, where the local models were combined to form a globally optimized model. Finally, the produced global models were further trained by each personalisation algorithm (in parallel). At every step, the performance of the models at each training stage were observed and documented. The experiments showed that, depending on the data and operation, the optimisation algorithms aided the federated models to better predict data on the local node. All of the models were tested against never-before seen data to provide a fair and accurate depiction of their performance.

These experimental findings reinforce the efficacy of the personalization module and the significance of employing Federated Learning techniques to optimize AI models. The insights gained from these experiments shed light on the impact of hyperparameters and data similarity on the performance of personalization modules, providing valuable information for future developments in Federated Learning and personalization techniques. By leveraging these insights, the optimization of AI models can be enhanced through personalized training and Federated Learning, enabling new applications and use cases for this technology.

The experiment results presented below were produced by applying the personalisation methods after the federated learning process for each model. For comparisson, the local training is also provided.

The observed results demonstrate that Active Learning, Knowledge Distillation and Local Memorisation can indeed be effective in optimising and personalising models. 

In particular, as it can be observed regarding the Animal Feed Cultivation Dataset, Active Learning and Knowledge Distillation are very effective in personalising FL models. Active Learning takes the lead in Table \ref{table:Animal Feed Cultivation Experiment Results [Target: CH4]}, while Knowledge Distillation outperforms the other methods in Table \ref{table:Animal Feed Cultivation Experiment Results [Target: Air Humidity]}.  This is expected as the leveraged models are good in capturing mid-dimentionality data distributions. This can be seen for all leveraged models. 
Nevertheless, there are some exceptions in Transformer case studies for MAE and RSME metrics in Table \ref{table:Animal Feed Cultivation Experiment Results [Target: Air Humidity]}. Given that the models were trained on various nodes, each having unique datasets, such results can be justified. The slightly varying data distributions that result from the data being collected in different areas are the cause of these variances in datasets. This can be also explained by the fact that temperature follows unique data distributions across different physical localities (e.g., different geographical locations) along with seasonality and periodicity changes. Even though conventional and temporal models (LSTM) can capture these changes, models like transformers have limited accumulation of the facts. This can be fixed by adding seasonality constraints to the models, which is out of the scope of this paper and would make the comparison biased.

Regarding the Smart Home Dataset, the majority of results exhibit promising performance improvements. More specifically, in  Table \ref{table:Smart Home Experiment Results [Target: User Behaviour]}, Knowledge Distillation has outperformed the other methods for all metrics. Nevertheless, in the case of DNN the FL model seems to have produced better results than the three Personalisation methods. This is expected as without customisation to adjust for the user behaviour variance the model is to weak to capture the data distribution. Models like the LSTM and the Transorment, handling both temporal and feature attention repsectively, seem more prone to produce more generalised resuts, which are further enhanced by personalisation. On the other hand, in  Table \ref{table:Smart Home Experiment Results [Target: Energy Consumption]}, Active Learning has produced better results for all metrics in Linear Regression case study, while the FL model outperforms the Personlaisation methods in the DNN and LSTM case studies. Similar to the results of the Animal Feed Cultivation Dataset, the Transformer case study remains an exception. 

Diverse results are also presented in the Dairy Product Sales Dataset. Table \ref{table:Dairy Product Sales Experiment Results [Target: Unit Sales]} demonstrates very positive results for Local Memorisation in Linear Regression as well as in DNN case studies for all metrics. On the other hand, Knowledge Distillation takes the lead in LSTM case studies, while Local Memorisation is favoured once again in the Transformer case studies.

When considering the overall results, it is evident that although there were a few outliers and difficulties, most of the findings show encouraging improvements. The case studies also highlight the importance of adjusting hyperparameters for particular model designs. In particular, Transformers require more research in this area since they provide a highly promising future for optimal applications.

\begin{table*}[!ht]
\caption{Animal Feed Cultivation Experiment Results [Target: CH4]}
\label{table:Animal Feed Cultivation Experiment Results [Target: CH4]}
\begin{adjustbox}{width=\textwidth}
\centering
\begin{tabular}{|ccccccccccccccccccccc|}
\hline
\multicolumn{21}{|c|}{\textbf{Mean Square Error (MSE)}}                                                                                                                                                                                                                                                                                                                                                                                                                                                                                                                                                                                                                                \\ \hline
\multicolumn{6}{|c|}{Linear Regression}                                                                                                                                                      & \multicolumn{5}{c|}{DNN}                                                                                                                                              & \multicolumn{5}{c|}{LSTM}                                                                                                                                             & \multicolumn{5}{c|}{Transformer}                                                                                                        \\
\hline
\end{tabular}
 \end{adjustbox}
\end{table*}

\begin{table*}[!ht]
\caption{Animal Feed Cultivation Experiment Results [Target: Air Humidity]}
\label{table:Animal Feed Cultivation Experiment Results [Target: Air Humidity]}
\begin{adjustbox}{width=\textwidth}
\centering
\begin{tabular}{|ccccccccccccccccccccc|}
\hline
\multicolumn{21}{|c|}{\textbf{Mean Square Error (MSE)}}                                                                                                                                                                                                                                                                                                                                                                                                                                                                                                                                                                                                                       \\ \hline
\multicolumn{6}{|c|}{Linear Regression}                                                                                                                                                      & \multicolumn{5}{c|}{DNN}                                                                                                                                     & \multicolumn{5}{c|}{LSTM}                                                                                                                                    & \multicolumn{5}{c|}{Transformer}                                                                                                                 \\
\hline
\end{tabular}
 \end{adjustbox}
\end{table*}

\begin{table*}[!ht]
\caption{Animal Feed Cultivation Experiment Results [Target: Air Temperature]}
\label{table:Animal Feed Cultivation Experiment Results [Target: Air Temperature]}
\begin{adjustbox}{width=\textwidth}
\centering
\begin{tabular}{|ccccccccccccccccccccc|}
\hline
\multicolumn{21}{|c|}{\textbf{Mean Square Error (MSE)}}                                                                                                                                                                                                                                                                                                                                                                                                                                                                                                                                                                                                              \\ \hline
\multicolumn{6}{|c|}{Linear Regression}                                                                                                                                                      & \multicolumn{5}{c|}{DNN}                                                                                                                                     & \multicolumn{5}{c|}{LSTM}                                                                                                                                    & \multicolumn{5}{c|}{Transformer}                                                                                                        \\ \hline
\multicolumn{1}{|c|}{}        & \multicolumn{1}{c|}{LC}     & \multicolumn{1}{c|}{FL}     & \multicolumn{1}{c|}{\textbf{KD}}     & \multicolumn{1}{c|}{AL}     & \multicolumn{1}{c|}{LM}     & \multicolumn{1}{c|}{LC}     & \multicolumn{1}{c|}{FL}     & \multicolumn{1}{c|}{\textbf{KD}}     & \multicolumn{1}{c|}{AL}     & \multicolumn{1}{c|}{LM}     & \multicolumn{1}{c|}{LC}     & \multicolumn{1}{c|}{FL}     & \multicolumn{1}{c|}{\textbf{KD}}     & \multicolumn{1}{c|}{AL}     & \multicolumn{1}{c|}{LM}     & \multicolumn{1}{c|}{\textbf{LC}}     & \multicolumn{1}{c|}{FL}     & \multicolumn{1}{c|}{KD}     & \multicolumn{1}{c|}{AL}     & LM     \\
\hline
\end{tabular}
 \end{adjustbox}
\end{table*}

\begin{table*}[!ht]
\caption{Smart Home Experiment Results [Target: User Behaviour]}
\label{table:Smart Home Experiment Results [Target: User Behaviour]}
\begin{adjustbox}{width=\textwidth}
\centering
\begin{tabular}{|ccccccccccccccccccccc|}
\hline
\multicolumn{21}{|c|}{\textbf{Mean Square Error (MSE)}}                                                                                                                                                                                                                                                                                                                                                                                                                                                                                                                                                                                                                       \\ \hline
\multicolumn{6}{|c|}{Linear Regression}                                                                                                                                                      & \multicolumn{5}{c|}{DNN}                                                                                                                                     & \multicolumn{5}{c|}{LSTM}                                                                                                                                             & \multicolumn{5}{c|}{Transformer}                                                                                                        \\
\hline
\end{tabular}
\end{adjustbox}
\end{table*}

\begin{table*}[h]
\caption{Smart Home Experiment Results [Target: Energy Consumption]}
\label{table:Smart Home Experiment Results [Target: Energy Consumption]}
\begin{adjustbox}{width=\textwidth}
\centering
\begin{tabular}{|ccccccccccccccccccccc|}
\hline
\multicolumn{21}{|c|}{\textbf{Mean Square Error (MSE)}}                                                                                                                                                                                                                                                                                                                                                                                                                                                                                                                                                                                                              \\ \hline
\multicolumn{6}{|c|}{Linear Regression}                                                                                                                                                      & \multicolumn{5}{c|}{DNN}                                                                                                                                     & \multicolumn{5}{c|}{LSTM}                                                                                                                                    & \multicolumn{5}{c|}{Transformer}                                                                                                        \\ \hline
\multicolumn{1}{|c|}{}        & \multicolumn{1}{c|}{LC}     & \multicolumn{1}{c|}{FL}     & \multicolumn{1}{c|}{KD}     & \multicolumn{1}{c|}{\textbf{AL}}     & \multicolumn{1}{c|}{LM}     & \multicolumn{1}{c|}{LC}     & \multicolumn{1}{c|}{\textbf{FL}}     & \multicolumn{1}{c|}{KD}     & \multicolumn{1}{c|}{AL}     & \multicolumn{1}{c|}{LM}     & \multicolumn{1}{c|}{LC}     & \multicolumn{1}{c|}{\textbf{FL}}     & \multicolumn{1}{c|}{KD}     & \multicolumn{1}{c|}{AL}     & \multicolumn{1}{c|}{LM}     & \multicolumn{1}{c|}{\textbf{LC}}     & \multicolumn{1}{c|}{FL}     & \multicolumn{1}{c|}{KD}     & \multicolumn{1}{c|}{AL}     & LM     \\
\hline
\end{tabular}
\end{adjustbox}
\end{table*}

\begin{table*}[h]
\caption{Dairy Product Sales Experiment Results [Target: Unit Sales]}
\label{table:Dairy Product Sales Experiment Results [Target: Unit Sales]}
\begin{adjustbox}{width=\textwidth}
\centering
\begin{tabular}{|ccccccccccccccccccccc|}
\hline
\multicolumn{21}{|c|}{\textbf{Mean Square Error (MSE)}}                                                                                                                                                                                                                                                                                                                                                                                                                                                                                                                                                                                                                       \\ \hline
\multicolumn{6}{|c|}{Linear Regression}                                                                                                                                                      & \multicolumn{5}{c|}{DNN}                                                                                                                                     & \multicolumn{5}{c|}{LSTM}                                                                                                                                    & \multicolumn{5}{c|}{Transformer}                                                                                                                 \\
\hline
\end{tabular}
 \end{adjustbox}
\end{table*}

\section{CONCLUSIONS}
\label{Conclusions}

A unique strategy that improves AI model customisation and optimisation is Personalisation, which enables each model to be specifically adapted to the requirements of each decentralised edge node. In data centre contexts, a distributed training module was found to be more effective than a centralised one, particularly when running on networks with low bandwidth or high latency. Three main methods of Personalisation were chosen and put into practice: local memorization, knowledge distillation, and active learning.
Choosing ambiguous data samples from a larger set of data and submitting them for labelling to a human expert is known as Active Learning. Training a smaller model to mimic the behaviour of a bigger, pre-trained model is known as Knowledge Distillation. By storing data locally on the edge node, Local Memorization helps the model remember important data samples that improve performance overall. It is easier to adapt the global AI model to each edge node's unique needs when Personalisation strategies like Knowledge Distillation, Active Learning, and Local Memorization are used.

In order to fully explore their potential in real-world scenarios within local and federated ecosystems, our study presented a Centralised federated Learning System that employed particular methodologies for Personalisation. This enabled the assessment of Knowledge Distillation, Active Learning, and Local Memorization in-depth within the framework of a machine learning system that protects privacy. Promising results were seen in the evaluated case studies, indicating the usefulness of Knowledge Distillation, Active Learning, and Local Memorization as tools to improve models that have already benefited from Federated Learning. The study's results hold great promise for the practical application of Knowledge Distillation, Active Learning, and Local Memorization to additional datasets, provided that the model and distiller parameters are calibrated appropriately. Furthermore, the results demonstrate that even in decentralised systems, personalised Federated Learning (FL) models provide improved predictive skills.

\section*{ACKNOWLEDGEMENT}
This project has received funding from the European Union’s Horizon Europe research and innovation programme under grant agreement No. 101135800 (RAIDO).

\bibliographystyle{IEEEtran}
\bibliography{bib}

\end{document}